\documentclass[10pt,journal,compsoc]{IEEEtran}
\usepackage[nocompress]{cite}
\usepackage[pdftex]{graphicx}
\graphicspath{{figures/}}
\DeclareGraphicsExtensions{.pdf,.jpeg,.png}
\usepackage{amsmath,bm}
\usepackage{algorithmic}
\usepackage{array}
\usepackage{url}
\usepackage[x11names]{xcolor}
\usepackage{tabulary,xfrac,enumitem}
\usepackage{booktabs}       
\usepackage{amsfonts}       
\usepackage{nicefrac}       
\usepackage{microtype}      
\usepackage{amssymb}
\usepackage{amsmath}
\usepackage{caption, subcaption}
\usepackage{multirow}
\usepackage{textcomp}
\usepackage{makecell}
\usepackage{colortbl}
\usepackage{pifont}
\usepackage{xspace}
\usepackage[ruled,noresetcount]{algorithm2e}
\usepackage[pagebackref=true,breaklinks=true,colorlinks,citecolor=citecolor,bookmarks=false]{hyperref}

\definecolor{citecolor}{RGB}{34,139,34}
\newcommand{\pub}[1]{\color{gray}{\tiny{[{#1}]}}}
\newcommand{\cmark}{\ding{51}\xspace}
\newcommand{\cmarkg}{\textcolor{lightgray}{\ding{51}}\xspace}
\newcommand{\xmark}{\ding{55}\xspace}
\newcommand{\xmarkg}{\textcolor{lightgray}{\ding{55}}\xspace}
\newcommand{\na}{\textcolor{lightgray}{-}}
\newcommand{\slice}[1]{\scalebox{0.8}{$\bm[{#1}\bm]$}}
\def \pzo {\phantom{0}}

\makeatletter
\newcommand{\thickhline}{%
    \noalign {\ifnum 0=`}\fi \hrule height 0.5pt
    \futurelet \reserved@a \@xhline
}
\makeatother

\begin{document}
\title{DS-Net++: Dynamic Weight Slicing for Efficient Inference in CNNs and Transformers}

\author{Changlin~Li,
        Guangrun~Wang,
        Bing~Wang,
        Xiaodan~Liang,
        Zhihui~Li,
        Xiaojun~Chang%
\IEEEcompsocitemizethanks{%
\IEEEcompsocthanksitem C. Li is with GORSE Lab, Dept. of DSAI, Monash University, Melbourne, Australia. Email: changlin.li@monash.edu.
\IEEEcompsocthanksitem G. Wang is with Oxford University. Email: wanggrun@gmail.com.
\IEEEcompsocthanksitem B. Wang is with Alibaba Group. Email: fengquan.wb@alibaba-inc.com.
\IEEEcompsocthanksitem X. Liang is with Sun Yat-sen University and DarkMatter AI Research. Email: xdliang328@gmail.com.
\IEEEcompsocthanksitem Z. Li is with Shandong Artificial Intelligence, Qilu University of Technology. Email: zhihuilics@gmail.com.
\IEEEcompsocthanksitem X. Chang is with RMIT University, Melbourne, Australia. Email: xiaojun.chang@rmit.edu.au}%
\thanks{Corresponding author: Xiaojun Chang.}%
}

\markboth{Work In Progress}%
{}

\IEEEtitleabstractindextext{%
\begin{abstract}
Dynamic networks have shown their promising capability in reducing theoretical computation complexity by adapting their architectures to the input during inference. However, their practical runtime usually lags behind the theoretical acceleration due to inefficient sparsity.
Here, we explore a hardware-efficient dynamic inference regime, named dynamic weight slicing. Instead of adaptively selecting important weight elements in a sparse way, we pre-define dense weight slices with different importance level by generalized residual learning. During inference, weights are progressively sliced beginning with the most important elements to less important ones to achieve different model capacity for inputs with diverse difficulty levels.
Based on this conception, we first present dynamic slimmable network (DS-Net) by input-dependently adjusting filter numbers of convolution neural networks. 
By extending dynamic weight slicing to multiple dimensions in both CNNs and transformers (e.g. kernel size, embedding dimension, number of heads, etc.), we further present dynamic slice-able network (DS-Net++). Both DS-Net and DS-Net++ adaptively slice a part of network parameters for inference while keeping it stored statically and contiguously in hardware to prevent the extra burden of sparse computation.
To ensure sub-network generality and routing fairness, we propose a disentangled two-stage optimization scheme. In Stage-I, a series of training techniques for dynamic supernet based on in-place bootstrapping (IB) and multi-view consistency (MvCo) are proposed to improve the supernet training efficacy. In Stage-II, sandwich gate sparsification (SGS) is proposed to assist the gate training.
Extensive experiments on 4 datasets and 3 different network architectures demonstrate our DS-Net and DS-Net++ consistently outperform their static counterparts as well as state-of-the-art static and dynamic model compression methods by a large margin (up to 6.6\%). Typically, DS-Net++ achieves 2-4$\times$ computation reduction and 1.62$\times$ real-world acceleration over MobileNet, ResNet-50 and Vision Transformer, with minimal accuracy drops (0.1-0.3\%) on ImageNet.
Code release: \url{https://github.com/changlin31/DS-Net}
\end{abstract}

\begin{IEEEkeywords}
Adaptive inference, dynamic networks, dynamic pruning, efficient inference, efficient transformer, vision transformer.
\end{IEEEkeywords}
}

\maketitle

\IEEEraisesectionheading{\section{Introduction}\label{sec:introduction}}
\IEEEPARstart{A}{s} deep neural networks are becoming deeper and wider to achieve better performance, there is an urgent need to explore efficient models for common mobile platforms, such as self-driving cars, smartphones, drones and robots. In recent years, many different approaches have been proposed to improve the inference efficiency of neural networks, including network pruning~\cite{Li2016PruningFF, liu2017NetworkSlimming, SoftFilterPruning, He2017ChannelPF, Liu2019MetaPruningML, luo2017thinet}, weight quantization~\cite{jacob2018quantization}, knowledge distillation~\cite{Ba2013DoDN, Romero2014FitNetsHF, Hinton2015DistillingTK}, manually and automatically designing of efficient networks~\cite{Tan2019EfficientNetRM,Sandler2018MobileNetV2IR,ZhangLPCZGS21,RenXCHLCW2020,bender2018understanding, ZhangLPCGS20, ChengZHDCLDG20, Guo2019SinglePO, li2019blockwisely, ZhangLPCS20} and dynamic inference~\cite{Bolukbasi2017AdaptiveNN, Huang2018MultiScaleDN, wang2018skipnet, veit2018AIG, Li2019ImprovedTF,hua2019channel, gao2018dynamic}. 

\begin{figure*}
\centering
{
\centering
\begin{subfigure}[t]{0.3\textwidth}
     \centering
    \includegraphics[width=\linewidth]{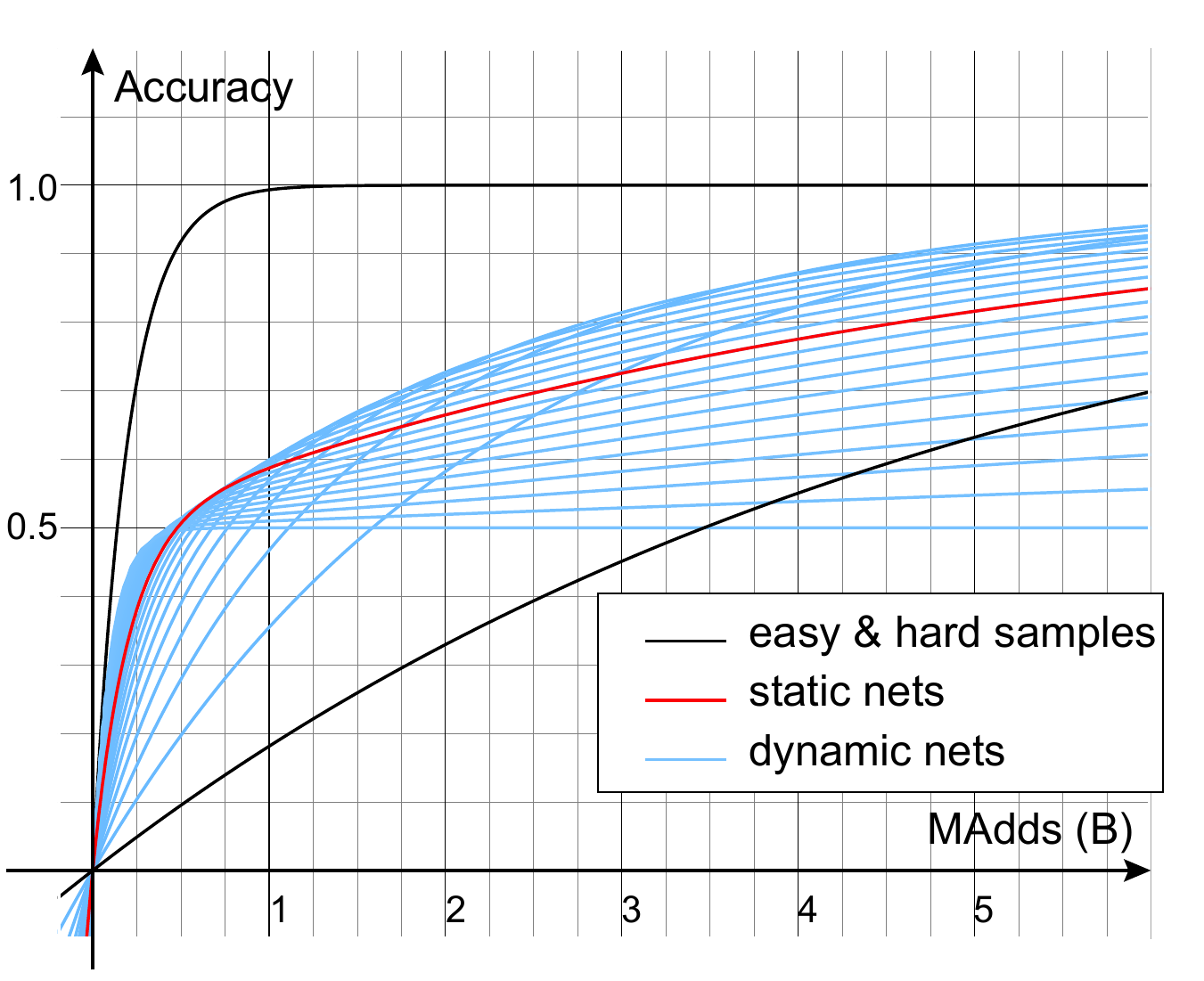}%
     \caption{}
\end{subfigure}
\hfil
\begin{subfigure}[t]{0.5\textwidth}
     \centering
    \includegraphics[width=\linewidth]{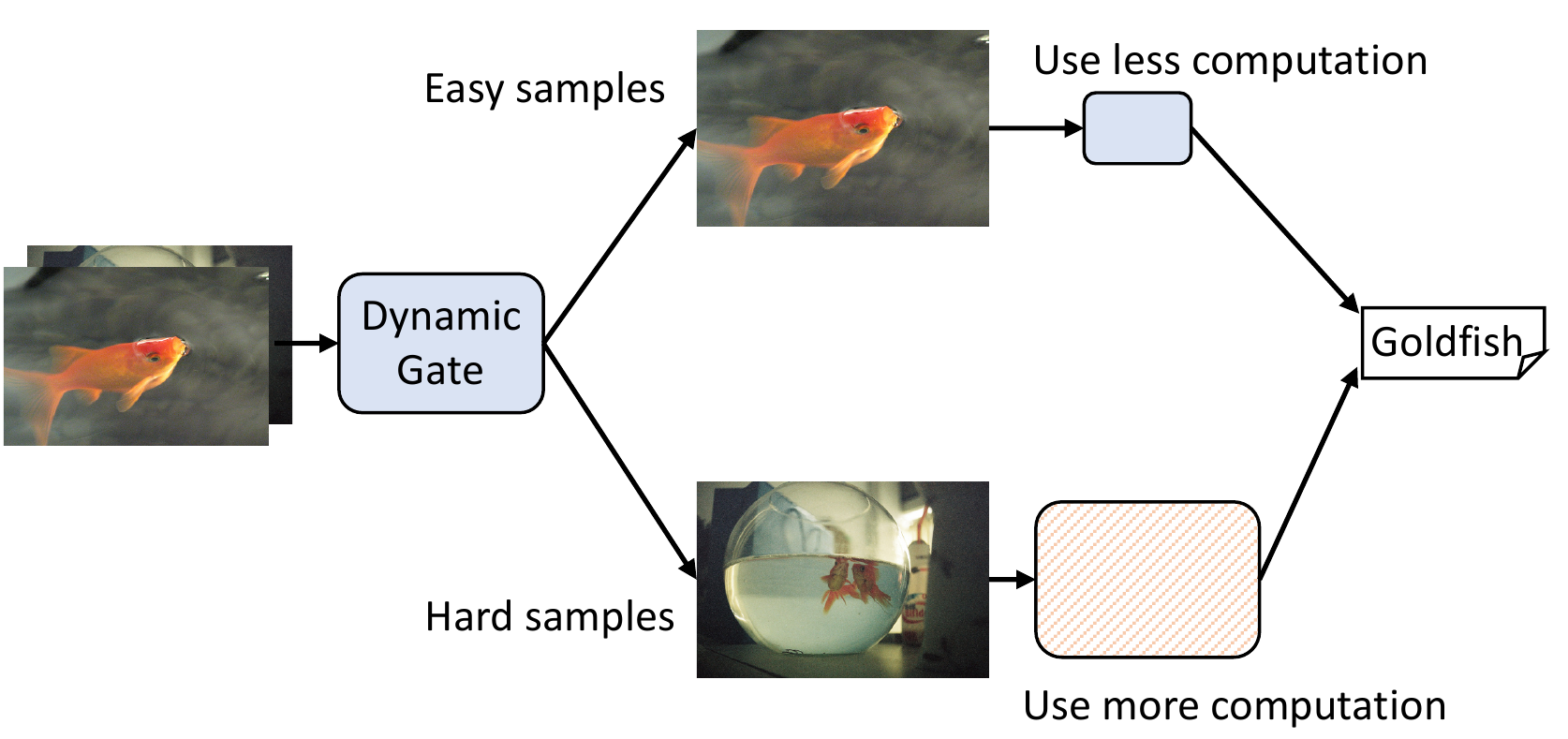}
     \caption{}
\end{subfigure}
}
\caption{
The motivation for designing dynamic networks to achieve efficient inference. 
\textbf{(a)} A simulation diagram of accuracy-complexity comparing a series of static networks (\textcolor{Red3}{red} curve) with 20 dynamic inference schemes (\textcolor{Blue3}{blue} curves) of different resource allocate proportion for easy and hard samples on a hypothetical classification task with evenly distributed easy and hard samples.
\textbf{(b)} Illustration of dynamic networks on efficient inference. Input images are routed to use different architectures regarding their classification difficulty.}\label{fig:motivation}
\end{figure*}

Among the above approaches, dynamic inference methods have attracted increasing attention because of their promising capability of reducing computational redundancy by automatically adjusting their architectures for different inputs (see Fig. \ref{fig:motivation} (b)), in contrast to static Neural Architecture Search (NAS) or Pruning methods that optimize the architecture for the whole dataset. A performance-complexity trade-off simulated with exponential functions is shown in Fig.~\ref{fig:motivation}~(a), the optimal solution of dynamic networks is superior to the static NAS or pruning solutions. Ideally, dynamic network routing can significantly improve model efficiency.

However, as point out in \cite{han2021dynamic}, the practical runtime of dynamic networks usually lags behind the theoretical complexity. For example, the networks with dynamic width, \textit{i.e.}, dynamic pruning methods~\cite{gao2018dynamic,hua2019channel,Chen2019YouLT}, unlike its orthogonal counterparts with dynamic depth, have never achieved actual acceleration in a real-world implementation. As natural extensions of network pruning, dynamic pruning methods predictively prune the convolution filters with regard to different input at runtime. The varying sparse patterns are incompatible with computation on hardware. Actually, many of them are implemented as zero masking or inefficient path indexing, resulting in a massive gap between the theoretical analysis and the practical acceleration. Similar issues also exist in pixel-wise dynamic inference methods \cite{ren2018sbnet, dong2017more,cao2019seernet,xie2020spatially}, sparse attention transformers \cite{2020sparsesinkhorn,haoyietal2021informer}, adaptive kernel CNNs \cite{dai2017deformable,Zhu2019DeformableCV,Gao2020DeformableKA}, etc.

To address the aforementioned issues in dynamic networks, we explore a hardware-efficient dynamic inference regime. 
Based on analysis of \textit{generalized residual learning} and a series of controlled experiments, we derive several practical guidelines for dynamic supernet design.
\textit{Dynamic weight slicing} scheme is accordingly proposed following the guidelines. Instead of predictively pruning convolution filters or adaptively selecting important weight elements in a sparse way, we pre-define dense weight slices with different importance level in a dynamic supernet by generalized residual learning. During inference, weights are progressively sliced beginning with the most important elements to less important ones to achieve different model capacity for inputs with diverse difficulty level.

Based on dynamic weight slicing regime, we first present Dynamic Slimmable Network (DS-Net) by adaptively adjusting filter numbers of convolution neural networks during inference with respect to different inputs. To avoid the extra burden on hardware caused by dynamic sparsity, filters are kept static and contiguous when adjusting the network width.
By extending dynamic weight slicing to multiple dimensions in both CNNs and transformers (e.g. kernel size, embedding dimension, number of heads, etc.), we further present dynamic slice-able network (DS-Net++). 
Dynamic routing is achieved via a \textit{double-headed dynamic gate} comprised of an attention head and a routing head with negligible extra computational cost. 

The training of dynamic networks is a highly entangled bilevel optimization problem. To ensure sub-network generality and the fairness of gate, a disentangled \textit{two-stage optimization scheme} is proposed to optimize the supernet and the gates separately. In \textit{Stage-I}, the dynamic supernet is optimized with a novel training method for weight-sharing networks, named \textit{in-place bootstrapping (IB)}. IB introduces a momentum supernet and trains the smaller sub-networks by predicting the output of the largest sub-network in the momentum supernet. Ensemble in-place bootstrapping, a variant of IB is further proposed to stabilize supernet training by encouraging the smallest sub-network to fit the output probability ensemble of larger sub-networks in the momentum supernet.
Advanced training techniques, hierarchical in-place bootstrapping (H-IB) and multi-view consistency (MvCo) are further proposed to improve the performance of DS-Net++. In H-IB scheme, intermediate supervision from teacher model is used to reduce the gap between sub-networks and ease the convergence hardship. MvCo scheme is performed by reducing the inconsistency among the results predicted from different augmented views of the same training sample using different sub-networks. 
In \textit{Stage-II}, to prevent dynamic gates from collapsing into static ones in the multi-objective optimization problem, a technique named \textit{sandwich gate sparsification (SGS)} is proposed to assist the gate training. During gate training, SGS identifies easy and hard samples online and further generates the ground truth label for the dynamic gates. 

Overall, our contributions are as follows:\begin{itemize}
\item{} We propose dynamic weight slicing scheme, achieving good hardware-efficiency by predictively slicing network parameters during inference with respect to different inputs, while keeping them stored statically and contiguously in hardware to prevent the extra burden of sparse computation.
\item{} We present DS-Net, by using dynamic weight slicing to adjust the filter number of CNNs. We propose a two-stage optimization scheme with IB, E-IB and SGS techniques for DS-Net.
\item{} By extending our dynamic weight slicing to more dimensions of CNNs, we present DS-CNN++. We additionally propose two advanced training schemes H-IB and MvCo for DS-CNN++. 
\item{} We further present DS-ViT++, by extending our dynamic weight slicing to multiple dimensions of vision transformers. To overcome the training difficulty when naively applying IB scheme on transformers, we re-designed an External Distillation scheme with MvCo for DS-ViT++ supernet.
\item{}Extensive experiments demonstrate our DS-Net and DS-Net++ outperform its static counterparts \cite{yu2019autoslim,Yu2019UniversallySN} as well as state-of-the-art static and dynamic model compression methods by a large margin (up to 6.6\%). Typically, DS-Net and DS-Net++ achieve 2-4$\times$ computation reduction and 1.62$\times$ real-world acceleration over MobileNet \cite{howard2017mobilenets}, ResNet-50 \cite{he2016deep} and ViT \cite{dosovitskiy2021vit} with minimal accuracy drops (0.1-0.3\%) on ImageNet.
\end{itemize}

\section{Related Work}\label{sec:related_work}
\subsection{Anytime Neural Networks}
Anytime neural networks are single networks that can execute with their sub-networks under different budget constraints, thus can be deployed instantly and adaptively in different application scenarios. Anytime neural networks have been studied in two orthogonal directions: networks with variable depth and variable width. Some recent works \cite{cai2019once,Yu2020BigNASSU,hou2020dynabert} also explore anytime neural networks in multiple dimensions, \textit{e.g.} depth, width, kernel size, \textit{etc.} 
\textit{Networks with variable depth} \cite{larsson2016fractalnet, Huang2018MultiScaleDN, hu2019learning, Li2019ImprovedTF} are first studied widely, benefiting from the naturally nested structure in depth dimension and residual connections in ResNet \cite{he2016deep} and DenseNet \cite{huang2017densely}. 
\textit{Network with variable width} was first studied in \cite{Lee2018AnytimeNP}. Recently, \textit{slimmable networks} \cite{Yu2019SlimmableNN, Yu2019UniversallySN} using \emph{switchable batch normalization} and \emph{in-place distillation} achieve better performance than their stand-alone counterparts in any width. In the field of Neural Architecture Search, AutoSlim \cite{yu2019autoslim}, Once-For-All\cite{cai2019once}, BigNAS \cite{Yu2020BigNASSU} and AutoFormer \cite{chen2021autoformer} are representative works that utilize a single anytime neural network to perform architecture search by evaluating its sub-networks.

\subsection{Dynamic Neural Networks}
Dynamic neural networks change their architectures based on the input data. For efficient inference, they reduce average inference cost by using different sub-networks adaptively \textit{w.r.t.} input difficulty. 

\textit{Dynamic CNNs.} Dynamic CNNs \cite{veit2018AIG,wang2018skipnet,li2020DynamicRouting,Yang2019CondConvCP} have been studied thoroughly in recent years.
\textit{Networks with dynamic depth} \cite{Bolukbasi2017AdaptiveNN, Huang2018MultiScaleDN, wang2018skipnet, veit2018AIG, Li2019ImprovedTF} achieve efficient inference in two ways, early exiting when shallower sub-networks have high classification confidence \cite{Bolukbasi2017AdaptiveNN, Huang2018MultiScaleDN, Li2019ImprovedTF}, or skipping residual blocks adaptively \cite{wang2018skipnet, veit2018AIG}. 
\textit{Dynamic pruning} methods~\cite{hua2019channel,gao2018dynamic,Chen2019YouLT} using a variable subset of convolution filters have been studied. Channel Gating Neural Network \cite{hua2019channel} and FBS \cite{gao2018dynamic} identify and skip the unimportant input channels at run-time. In GaterNet \cite{Chen2019YouLT}, a separate gater network is used to predictively select the filters of the main network. Please refer to \cite{han2021dynamic} for a more comprehensive review of dynamic neural networks.

\textit{Dynamic Transformer.}
Dynamic networks based on Transformer \cite{Vaswani2017Attention_nips} and BERT \cite{Devlin2019pretraining_naacl} have been proved effective in Neural Language Processing (NLP) Tasks. Most of the works focus on developing transformers with dynamic depth by early exiting \cite{zhou_bert_2020,liu_fastbert_2020,xin_deebert_2020,schwartz_right_2020} or skipping layers in blocks by halting scores \cite{dehghani_universal_2019,elbayad_depth-adaptive_2020}. Switch Transformer \cite{2021arXiv210103961F} uses Mixture-of-Expert (MoE) routing algorithm to switch between different feed forward network (FFN) layers based on the input.
Recently, the exciting break-through in vision transformers led by ViT \cite{dosovitskiy2021vit} and DETR \cite{carion2020detr} makes \textit{dynamic vision transformers} a promising research topic.
DVT \cite{wang2021not} automatically configures a decent token number conditioned on each image for high computational efficiency by early exiting from a cascade of transformers with increasing token numbers.
DynamicVit \cite{rao2021dynamicvit} uses binary gating module to selectively abandon less informative tokens conditioned on features produced by the previous layer.
V-MoE \cite{riquelme2021scaling} replaces a subset of the dense feedforward layers in ViT with Mixture-of-Experts (MoEs).

\subsection{Static Neural Architecture Optimization}
\textit{Weight sharing NAS.} Weight sharing NAS methods integrate the whole search space of NAS into a weight sharing supernet and optimize network architecture by pursuing the best-performing sub-networks. These methods can be roughly divided into two categories: \textit{jointly optimized methods} \cite{Liu2018DARTSDA, Cai2018ProxylessNASDN, Wu2018FBNetHE}, in which the weight of the supernet is jointly trained with the architecture routing agent (typically a simple learnable factor for each candidate route); and \textit{one-shot methods} \cite{brock2017smash, akimoto2019adaptive, bender2018understanding, Guo2019SinglePO, li2019blockwisely,cai2019once,li2021bossnas}, in which the training of the supernet parameters and architecture routing agent are disentangled. After fair and sufficient training, the agent is optimized with the weights of supernet frozen.

\section{Hardware-Efficient Dynamic Inference Scheme}\label{sec:methodology1}

\begin{figure*}[t]
    \centering
    {
    \centering
    \begin{subfigure}[t]{0.19\textwidth}
         \centering
        \includegraphics[height=3cm]{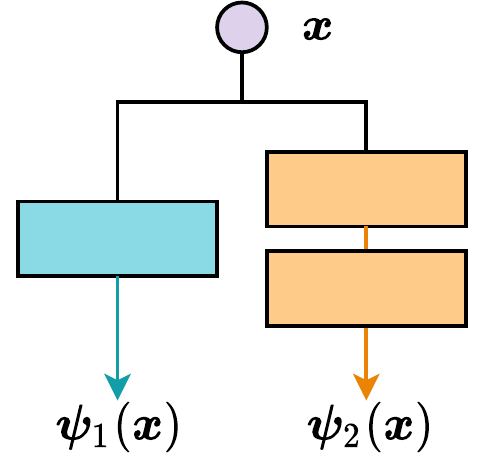}%
         \caption{Stand alone.}
     \end{subfigure}
     \hfill
    \begin{subfigure}[t]{0.19\textwidth}
         \centering
        \includegraphics[height=3cm]{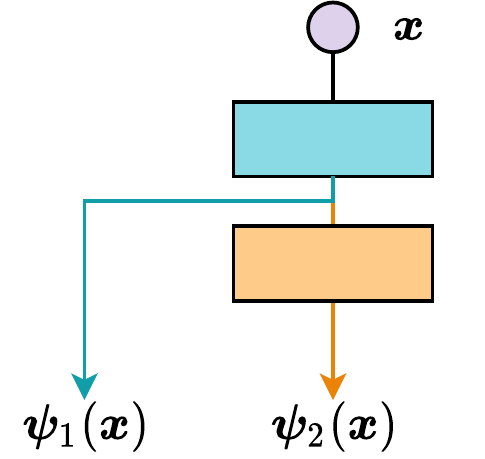}%
         \caption{Naive.}
     \end{subfigure}
     \hfill
     \begin{subfigure}[t]{0.19\textwidth}
         \centering
        \includegraphics[height=3cm]{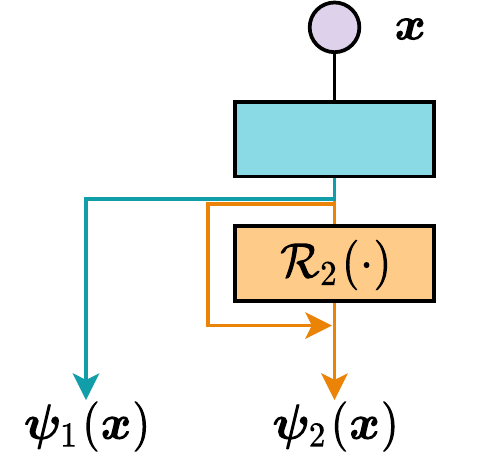}%
         \caption{Residual.}
     \end{subfigure}
     \hfill
     \begin{subfigure}[t]{0.19\textwidth}
         \centering
        \includegraphics[height=3cm]{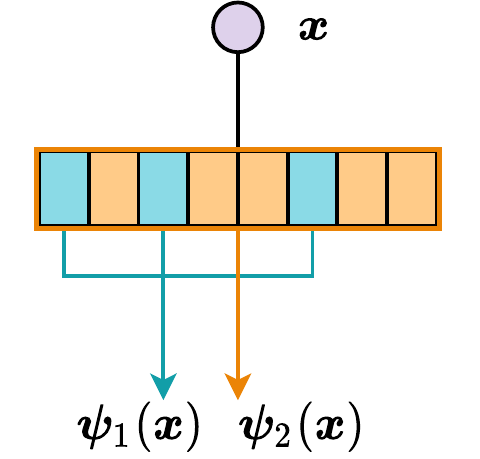}%
         \caption{Weight Nesting.}
     \end{subfigure}
     \hfill
     \begin{subfigure}[t]{0.19\textwidth}
         \centering
        \includegraphics[height=3cm]{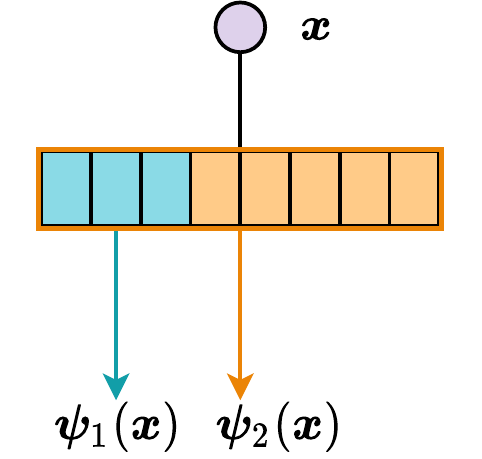}%
         \caption{Weight Slicing.}
     \end{subfigure}
    }
    \caption{Illustrations of generalized residual learning, weight nesting and weight slicing.}
    \label{fig:weight_slicing}
\end{figure*}

In this section, we present our hardware-efficient dynamic inference scheme in 3 steps. First, we derive dynamic weight slicing scheme with theoretical basis. Then, we apply dynamic weight slicing and present our dynamic slice-able network. Finally, we design an efficient routing agent.

\textit{Notations.} 
We denote tensors with fixed shapes (such as supernet parameters) using bold upper case letters (\textit{e.g.} $\textbf W$) and tensors with varying shapes (such as feature maps in dynamic networks) using bold upper case calligraphic letters (\textit{e.g.} $\bm{\mathcal{X}}$). Scalars are denoted with regular letters (\textit{e.g.} $d$, $D$).

\subsection{Dynamic Inference with Dynamic Supernet and Routing Agent}
We present a general dynamic inference scheme by a dynamic neural network $\mathcal{Q}(\textbf{W}, \bm\theta)$, consisting of a \textit{dynamic supernet} $\Psi(\textbf{W})$ with parameters $\textbf{W}$ undertaking the main task and a \textit{routing agent} $\Phi(\bm\theta)$, parameterized by $\bm\theta$, which predicts the optimal route in the dynamic supernet for each input.

\textit{Dynamic Supernet.} Formally, a dynamic supernet $\Psi(\textbf{W})$ is a weight-sharing network parameterized by $\textbf{W}$, which can execute with any of the $|t|$ different sub-networks $\{\mathcal{\bm\psi}_t(\textbf{w}_t)\}$, $\textbf{w}_t \subset \textbf{W}$, when given a routing signal. To achieve efficient inference, the supernet should contain sub-networks with different complexities, which will be selected during inference adjusting to the input difficulty. For simplicity, we denote sub-networks in order of complexity, with $\bm\psi_1$ being the smallest sub-network, and $\bm\psi_{|t|}$ being the largest sub-network.

\textit{Routing Agent.} A routing agent $\Phi(\bm\theta)$ is a light-weight module, predicting the states $\{\phi_t\}$ for each of the $|t|$ executable sub-networks in the dynamic supernet $\Psi$. The states can be represented as a vector $\bm\phi\in\{0,1\}^{|t|}$ of elements that are \textit{one-hot}, \textit{i.e.} $\sum_{t=0}^{|t|}\phi_t = 1$.

Then, the dynamic neural network is simply implemented as dot product of the routing agent and the dynamic supernet:
\begin{equation}
\label{eqn:dyn_net}
\begin{aligned}
    \mathcal{Q}(\textbf{W}, \bm\theta; \bm{x}) &= \Phi(\bm\theta; \bm{x}) \cdot \Psi(\textbf{W};\bm{x})\\
    &= \bm\psi^{\bm\phi(\bm\theta;\bm{x})}(\textbf{W};\bm{x}).
\end{aligned}
\end{equation}
where $\bm\psi^{\bm\phi}$ denotes the route selected by the indicator $\bm\phi$.

\subsection{Practical Guidelines for Dynamic Supernet Design}
\textbf{GUIDELINE I: Generalized Residual Learning Guarantees Lossless Weight Sharing.}
Considering $\mathcal{H}(\bm{x})$ as an underlying mapping to be fit by a network module $\mathcal{F}(\bm{x})$, with $\bm{x}$ denoting the input, \textbf{residual learning} \cite{he2016deep} is defined as learning a function $\mathcal{R}(\bm{x}) := \mathcal{H}(\bm{x}) - \bm{x}$, which makes the original function $\mathcal{F}(\bm{x}) = \bm{x} + \mathcal{R}(\bm{x})$. Here, we generalize the residual learning to supernets. Let us consider $\mathcal{H}(\bm{x})$ as the underlying mapping to be fit by each of the $|t|$ paths $\{\bm\psi_t\}$ of a supernet module $\Phi(\bm{x})$. Instead of learning each $\bm\psi_t$ separately to mimic $\mathcal{H}$, \textbf{generalized residual learning} makes each path (except for the first one) approximate the residual function of the previous path:
\begin{equation}
    \mathcal{R}_t(\bm{x}) := \mathcal{H}(\bm{x}) - \bm\psi_{t-1}(\bm{x}), ~~~~1<t\leq|t|.
\end{equation}
Thus, the original function of each path is split into a core module and $t$ residuals:
\begin{equation}
\label{eqn:residual}
    \bm\psi_{t}(\bm{x}) = \underbrace{\bm\psi_1(\bm{x})}_{\text{core}} + \sum\limits_{i=1}^{t}\underbrace{\mathcal{R}_{i}(\bm{x})}_{\text{residuals}}, ~~~~1<t\leq|t|.
\end{equation}
The smaller paths can always be achieved by learning residuals in larger paths to 0.

\textit{A Case Study: Dynamic Depth.}
For instance, a supernet module, consisting of two sequential blocks $f$ and $g$, has two paths $\bm\psi_1$ and $\bm\psi_2$. Generally, the supernet module has 3 possible scheme:
\begin{itemize}
\item{} \textit{Stand alone.} $\bm\psi_1$ and $\bm\psi_2$ are two separate paths without weight sharing:
\begin{equation}
\left\{
\begin{aligned}
    &\bm\psi_1(\bm{x}) = f_a(\bm{x}),\\
    &\bm\psi_2(\bm{x}) = g(f_b(\bm{x}))
\end{aligned}
\right.
\end{equation}
where $f_a$ and $f_b$ are block $f$ with different parameters;
\item{} \textit{Naive Weight Sharing.} $\bm\psi_1$ and $\bm\psi_2$ share part of their weights but do not form an additive relation:
\begin{equation}
\label{eqn:naivews}
\left\{
\begin{aligned}
    &\bm\psi_1(\bm{x}) = f(\bm{x}),\\
    &\bm\psi_2(\bm{x}) = g(\bm\psi_1(\bm{x}));
\end{aligned}
\right.
\end{equation}
\item{} \textit{Generalized Residual Learning.} $\bm\psi_1$ and $\bm\psi_2$ satisfy Eq. \eqref{eqn:residual}:
\begin{equation}
\label{eqn:residualdepth}
\left\{
\begin{aligned}
    &\bm\psi_1(\bm{x}) = f(\bm{x}),\\
    &\bm\psi_2(\bm{x}) = \bm\psi_1(\bm{x}) + g(\bm\psi_1(\bm{x})).
\end{aligned}
\right.
\end{equation}
\end{itemize}
Here, $\mathcal{R}_2(\bm{x}) := g(\bm\psi_1(\bm{x}))$ is the residual function to fit the underlying residual mapping $\big(\mathcal{H}(\bm{x}) - \bm\psi_1(\bm{x})\big)$. Fig. \ref{fig:weight_slicing} (a)-(c) provide simple illustrations of these 3 schemes. Here we show by experiments that learning $\bm\psi_1$ and $\mathcal{R}_2$ is easier than learning $\bm\psi_1$ and $\bm\psi_2$ separately. We replace the $3$-rd stage of MobileNet V1 with a supernet block that can execute with two routes: 1 layer or 6 layers.
The accuracy of the two routes in these three schemes, after 50,000 steps training on ImageNet, are shown in Table \ref{tab:residual}. Naive weight sharing leads to remarkable accuracy drop (-1.3\% in $\bm\psi_1$) in this simple supernet with only two paths. Thus, networks without skip connections, such as MobileNet V1 \cite{howard2017mobilenets}, is not suitable for dynamic depth. By adding a single skip connection from layer 1 to layer 6, generalized residual learning achieves better performance than stand alone, which proves that generalized residual learning guarantees lossless weight sharing.

\begin{table}[t]

\renewcommand{\arraystretch}{1.3}
    \centering
    \caption{Accuracy comparison of dynamic depth supernet with only 2 paths on MobileNet V1 \cite{howard2017mobilenets}. Naive weight sharing (\textit{Naive}) leads to accuracy drop in both of the sub-networks, while generalized residual learning (\textit{Residual}) guarantees lossless weight sharing. }
    \label{tab:residual}
    \setlength{\tabcolsep}{12pt}
    \begin{tabular}{l|c|c|c}
        \toprule
        Scheme              & Stand alone   & Naive     & Residual\\
        \hline
        $\bm\psi_1$ Acc.    & 47.4          & 46.1 (-1.3\%)      &\textbf{48.3 (+0.9\%)}\\
        $\bm\psi_2$ Acc.    & 50.7          & 50.0 (-0.7\%)      &\textbf{53.2 (+2.5\%)}\\
        \bottomrule
    \end{tabular}
\end{table}

\begin{figure*}[!t]
    \centering
    \includegraphics[width=0.9\linewidth]{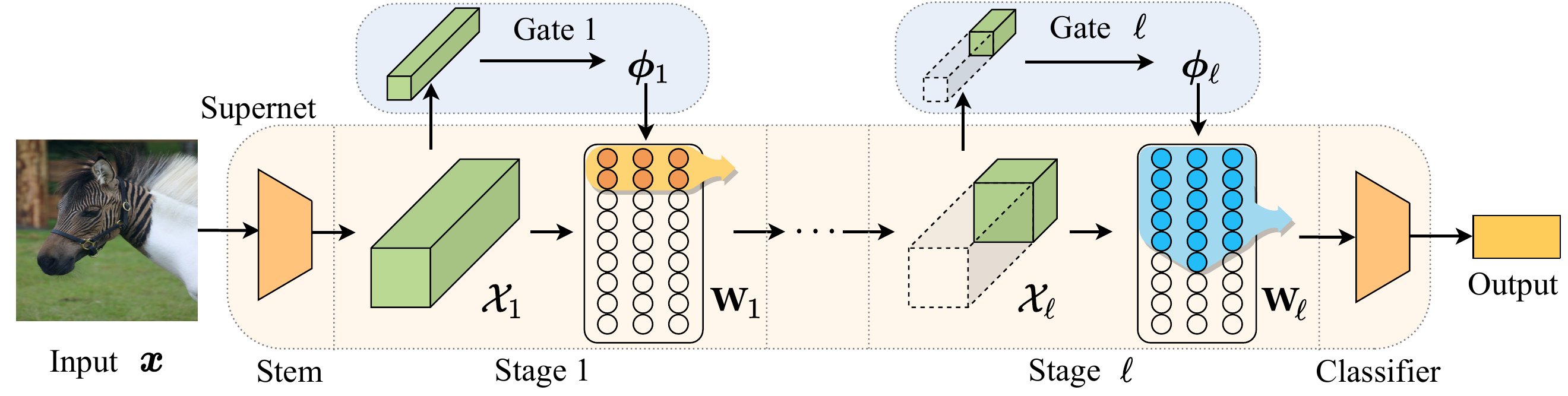}
    \caption{Architecture of dynamic slice-able convolution neural networks (DS-CNN++). The architecture of each supernet stage is adjusted adaptively by the routing signal $\bm\phi$ predicted by the gates. For simplicity, only channel dimension is showned in this figure, see Fig. \ref{fig:dynamic_kernel} for the illustration of dynamic slicing in spatial dimension.}\label{fig:dscnn}
\end{figure*}

\textit{Weight Nesting.}
Generalized residual learning of different layers or branches can be easily achieved by adding skip connections. However, too many skip connections increase network fragmentation, which can further reduce the degree of parallelism \cite{ma2018shufflenetv2}. Here, we further study generalized residual learning in a single weight layer.
Without loss of generality, we consider $\bm\psi_1(\bm{x}, \textbf{w}_1)$ and $\bm\psi_2(\bm{x}, \textbf{w}_2)$ being the first two of $|t|$ paths of a single fully-connected super layer $\Psi(\bm{x}, \textbf{W})$, approximating the same underlying mapping $\mathcal{H}(\bm{x})$. Cases with higher dimensional parameters, \textit{e.g.} convolutions, can be easily generalized.
Following generalized residual learning, instead of learning the two mappings separately, we nest $\textbf{w}_1$ into $\textbf{w}_2$ (see Fig. \ref{fig:weight_slicing} (d)),
and let $\bm\psi_2$ approximate the residual function following Eq. \eqref{eqn:residual}. We denote $\textbf{w}_t$ by the association of a mask $\bm{m}_t \in \{0, 1\}^d$ and supernet weight $\textbf{W} \in \mathbb{R}^d$. As $\textbf{w}_1\subset\textbf{w}_2$, the residual mask is simply implemented as $\bm{m}_{\mathcal{R}2}:=\bm{m}_2-\bm{m}_1 \in \{0, 1\}^d$. The generalized residual learning in weight nesting scheme are as follows:
\begin{equation}
\label{eqn:weightnesting}
\left\{
\begin{aligned}
    &\bm\psi_1(\bm{x}) = (\bm{m}_1\cdot\textbf{W})\bm{x},\\
    &\bm\psi_2(\bm{x}) = \bm\psi_1(\bm{x}) + (\bm{m}_{\mathcal{R}2}\cdot\textbf{W})\bm{x}.
\end{aligned}
\right.
\end{equation}

\begin{table}[!t]
\renewcommand{\arraystretch}{1.3}
 \caption{Latency comparison of ResNet-50 with 25\% channels (on GeForce RTX 2080 Ti).
Both \textit{masking} and \textit{indexing} lead to inefficient computation waste, while \textit{slicing} achieves comparable acceleration with \textit{ideal} (the individual ResNet-50 0.25$\times$).}
 \label{tab:ds_vs_dp}
 \centering
 \footnotesize
 \setlength{\tabcolsep}{5pt}
 \begin{tabular}{l|ccccc}
  \toprule
  Method  & Full & Masking & Indexing & Slicing & Ideal \\
  \hline
  Latency & 12.2 ms & 12.4ms & 16.6 ms & 7.9 ms  & 7.2 ms\\
  \bottomrule
 \end{tabular}
\end{table}

\textbf{GUIDELINE II: Weight Slicing Minimizes Sparse Inference Overhead.}
In previous pruning and dynamic pruning methods, the convolution filters are pruned irregularly, forming a sparse pattern. In static pruning methods \cite{SoftFilterPruning,luo2017thinet,Liu2019MetaPruningML,li2020eagleeye}, the sparse filters are merged to a dense one after pruning. However, in dynamic pruning methods \cite{hua2019channel,gao2018dynamic,Chen2019YouLT}, the filters can not be merged as they are changing input-dependently during inference. More specifically, these methods predictively generate the irregular pruning mask $\bm m$ and apply to the supernet weight $\textbf{W}$. To minimize the computation waste in sparse inference, the weights of sub-networks should be dense and regular.

\textit{Weight Slicing.} Based on weight nesting, we further achieve dense dynamic inference by weight slicing. In this scheme, weights in sub-networks are forced to be dense and regular. We consider a weight nesting fully-connected super layer $\Psi(\bm{x}, \textbf{W})$ with sub-networks $\bm\psi_1$ and $\bm\psi_2$ satisfying Eq. \eqref{eqn:weightnesting}. In weight slicing scheme, if we treat parameters as sequences, then $\textbf{w}_1$ is a contiguous subsequence of $\textbf{w}_2$ (see Fig. \ref{fig:weight_slicing} (e)). Without loss of generality, we assume all the subsequence begin from the first index of $\textbf{W}$. Let $k_t$ be the end index of $\textbf{w}_t$ in $\textbf{W}$, the supernet module is thus a single layer executing with different slice of weight:
\begin{equation}\label{eqn:weight_slicing}
\left\{
\begin{aligned}
    &\bm\psi_1(\bm{x}) = \textbf{W}\slice{:k_1}~\bm{x}, \\
    &\bm\psi_{2}(\bm{x}) = \bm\psi_1(\bm{x}) + \textbf{W}\slice{k_1:k_2}\bm{x},
\end{aligned}
\right.
\end{equation}
where $\slice{s:e}$ denotes a slicing operation from index $s$ to $e$, in python-like style. Table \ref{tab:ds_vs_dp} shows the speed comparison of different sparse inference schemes. Compared to weight masking and indexing, weight slicing minimizes sparse inference overhead.

By controlling the slicing index with an input-dependent routing signal $\bm\phi = \Phi(\bm{x})$, we have the hardware-efficient \textit{dynamic weight slicing} scheme.
In our conference version \cite{Li2021DynamicSN}, we have explored dynamic slicing in CNN filter numbers and presented dynamic slimmable network (DS-Net).
In the following sections, we generalize dynamic weight slicing to multiple dimensions of CNNs and transformers, and accordingly present dynamic slice-able network (DS-Net++).

\begin{figure}[t]
    \centering
    {
    \centering
    \begin{subfigure}[t]{0.49\linewidth}
         \centering
        \includegraphics[width=\linewidth]{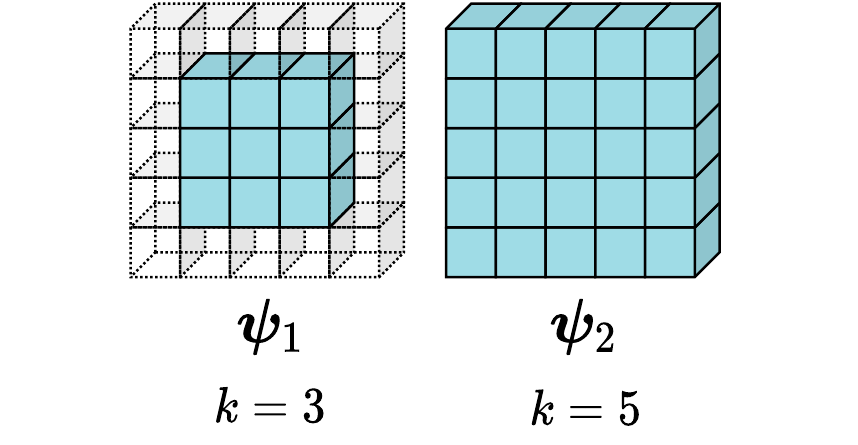}%
         \caption{Kernel size.}
     \end{subfigure}
    \begin{subfigure}[t]{0.49\linewidth}
         \centering
        \includegraphics[width=\linewidth]{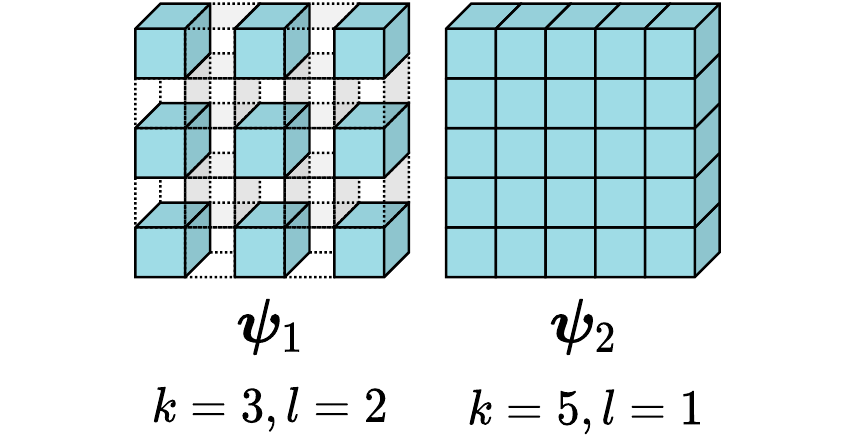}%
         \caption{Dilation.}
     \end{subfigure}
         }
    \caption{Illustrations of weight slicing in spatial dimension of convolution filters.}
    \label{fig:dynamic_kernel}
\end{figure}
\subsection{Dynamic Slice-able Convolution Neural Networks}
In this section, we present dynamic slice-able convolution neural networks (DS-CNN++). As shown in Fig. \ref{fig:dscnn}, DS-CNN++ achieves dynamic inference with dynamic weight slicing on multiple dimensions of CNNs.

\begin{figure*}[t]
    \centering
    \includegraphics[width=0.8\linewidth]{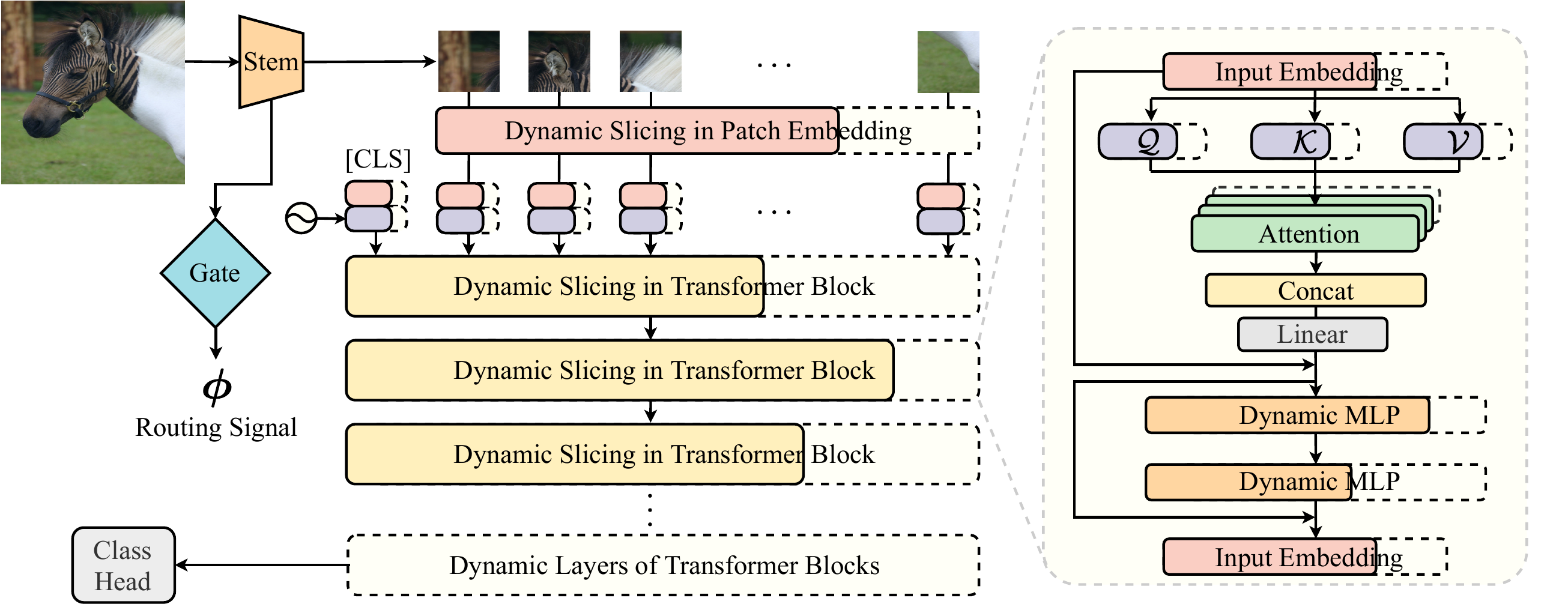}
    \caption{Architecture of dynamic slice-able vision transformer (DS-ViT++). Embedding dimension, Q-K-V dimension, head number, MLP ratio and layer number are predictively adjusted by routing signal $\bm\phi$, with respect to different inputs.}\label{fig:dsvit}
\end{figure*}

\textit{Dynamic Slicing in Convolution.} 
Let $\textbf{W}\in\mathbb{R}^{C_{\mathcal{O}}\times C_{\mathcal{I}} \times K\times K}$ denote the parameters of a convolutional super layer in weight slicing scheme, with $C_{\mathcal{I}}$, $C_{\mathcal{O}}$, $K$ denoting the input channel number, output channel number and kernel size.
Similarly, in group convolution, we have parameter $\textbf{W}\in\mathbb{R}^{N\times G_{\mathcal{O}}\times G_{\mathcal{I}} \times K\times K}$, with $N$, $G_{\mathcal{I}}$, $G_{\mathcal{O}}$ being the group number, input channel number and output channel number per group, respectively.
Given an elastic input with certain shape $\bm{\mathcal{X}}\in\mathbb{R}^{c_{\mathcal{I}}\times H\times W}$ ($c_{\mathcal{I}}\leq C_{\mathcal{I}}$), produced by the previous slimmable layer, a dynamic slice-able convolution $\Psi_{conv}(\bm{\mathcal{X}})$ can execute by any path $\bm\psi_{conv}(\bm{\mathcal{X}})$ with weight slices in multiple dimensions, adjusting to the routing signal $\bm\phi$.
\begin{itemize}
\item{} \textit{Filter number.} Similar to output dimension in fully-connected super layers, the filter number can be dynamically sliced as:
\begin{equation}
    \bm\psi_{conv}(\bm{\mathcal{X}}) = \textbf{W}\slice{:c^{\bm\phi}_{\mathcal{O}}, :c_{\mathcal{I}}}\star\bm{\mathcal{X}},
\end{equation}
where $\star$ denotes convolution and $c^{\bm\phi}_{\mathcal{O}}:= \bm\phi \cdot \{c_{\mathcal{O}}\}$ is the selected output channel, with $(\cdot)^{\bm\phi}$ denoting that the slicing index is dynamically controlled by the routing signal of current layer.
\item{} \textit{Group number and filter number per group.}
For slice-able group convolution, we have two schemes to reduce the filter number: dynamic group number or dynamic filter number per group. In the first scheme, the input is reshaped into $n\times G_{\mathcal{I}}\times H\times W$. While in the second scheme, the input is reshaped into $N\times g_{\mathcal{I}}\times H\times W$. These two schemes can be formulated as:
\begin{equation}
\left\{
\begin{aligned}
    &\bm\psi_{conv}(\bm{\mathcal{X}}) = \text{Concat}(\textbf{W}\slice{i}\star\bm{\mathcal{X}}\slice{i})_{i = 0}^{n};\\
    &\bm\psi_{conv}(\bm{\mathcal{X}}) = \text{Concat}(\textbf{W}\slice{i, :g^{\bm\phi}_{\mathcal{O}}, :g_{\mathcal{I}}}\star\bm{\mathcal{X}}\slice{i})_{i = 0}^{N}.
\end{aligned}
\right.
\end{equation}
Among these two schemes, the output filter number of the first scheme is determined only by the input shape, i.e. $c_{\mathcal{O}} = G_{\mathcal{O}} * n = G_{\mathcal{O}} *c_{\mathcal{I}}/G_{\mathcal{I}}$, while the output filter number of the second scheme can be determined by routing signal $\bm\phi$.
\item{} \textit{Kernel size.} As shown in Fig. \ref{fig:dynamic_kernel} (a), in spatial dimension, the filters can be sliced into different kernel sizes:
\begin{equation}
    \bm\psi_{conv}(\bm{\mathcal{X}}) = \textbf{W}\slice{:, :c_{\mathcal{I}}, k_0^{\bm\phi}:k_0^{\bm\phi}+k^{\bm\phi}, k_0^{\bm\phi}:k_0^{\bm\phi}+k^{\bm\phi}}\star\bm{\mathcal{X}}.
\end{equation}
\item{} \textit{Dilation.} Slicing operation with interval larger than one is also hardware efficient in modern deep learning framework, which makes dynamic dilation a possible scheme (see Fig. \ref{fig:dynamic_kernel} (b)). Let $l$ denote the dilation, and $\slice{::l}$ denote slicing with interval $l$, a convolution with dynamic dilation $l$ can be achieved by the following objective function:
\begin{equation}
    \bm\psi_{conv}(\bm{\mathcal{X}}) = \textbf{W}\slice{:, :c_{\mathcal{I}}, ::l^{\bm\phi}, ::l^{\bm\phi}}\star\bm{\mathcal{X}}.
\end{equation}
\item{} \textit{Padding.} By directly slicing the padded input feature map $\bm{\mathcal{X}}$, different padding size and output resolution can be achieved. Let $p$ denote the difference between the selected padding size and maximum padding size on a single side, dynamic padding can be achieved by the following objective function:
\begin{equation}
    \bm\psi_{conv}(\bm{\mathcal{X}}) = \textbf{W}\star\bm{\mathcal{X}}\slice{:, p^{\bm\phi}:-p^{\bm\phi}, p^{\bm\phi}:-p^{\bm\phi}}.
\end{equation}
\end{itemize}

\textit{Split Batch Normalization with Re-calibration.}
Batch Normalization plays a vital role in stabilizing the dynamics in CNNs. Let $\bm\gamma \in \mathbb{R}^C$, $\bm\beta \in \mathbb{R}^C$ denote the scaling factors of a batch normalization super layer. To eliminate the distribution shift caused by the switching route, we calculate mean $\bm\mu_t$ and standard deviation $\bm\sigma_t$ of the input separately for each route $t$ following \cite{xie2019adversarial,Yu2019SlimmableNN}. After training, we re-calibrate $\bm\mu_t$ and $\bm\sigma_t$ for each path by cumulative average over the whole training set following \cite{Yu2019UniversallySN}, which typically takes 2 minutes on 8 GPUs for each path. Given an input $\bm{\mathcal{X}} \in \mathbb{R}^{c\times H\times W}$ ($c\leq C$) from route $t$, split batch normalization with re-calibration (Re-SBN) is formulated as follows:
\begin{equation}
    \bm\psi_{bn}(\bm{\mathcal{X}}) = \bm\gamma\slice{:c}\cdot\left(\frac{\bm{\mathcal{X}}-\bm\mu_t}{\bm\sigma_t}\right) + \bm\beta\slice{:c}.
\end{equation}

\subsection{Dynamic Slice-able Vision Transformer}

\begin{table}[t]

    \centering
    \caption{Complexity comparison of different patch embedding design. ``c'': convolution; ``ds'': Depthwise Separable block \cite{howard2017mobilenets}.}
    \begin{tabular}{l|c|c|c|c|c}
    \toprule
    Source & blocks & dim & kernel & stride & MAdds\\
    \midrule
         ViT/16 \cite{dosovitskiy2021vit} &c  & - & 16      &   16    & 5.5G \\
         LV-ViT \cite{jiang2021all}       &[c, c, c, c]    & 64 & [7,3,3,8] & [2,1,1,8] & 6.8G \\
         Ours                             &[c, ds, ds, c]    & 24 & [7,3,3,8] & [2,1,1,8] &  5.6G     \\
    \bottomrule
    \end{tabular}
    \label{tab:my_label}
\end{table}

In this section, we present dynamic slice-able vision transformer (DS-ViT++). As shown in Fig. \ref{fig:dsvit}, DS-ViT++ dynamically slice the parameters on multiple dimensions of vision transformers.

\textit{Efficient stem with depthwise separable convolution.}
In DS-CNN++, to ease the prediction of routing decision, the dynamic gate is inserted at intermediate layer and takes feature maps rather than the original image as the input. However, in vision transformer \cite{dosovitskiy2021vit}, the first layer, patch embedding, controls the embedding dimension of all the subsequent transformer blocks. To endow the transformer blocks with the ability of dynamically adjusting embedding dimension, we insert the gate before patch embedding layer and add a light weight stem before the gate to ease its prediction (see Fig. \ref{fig:dsvit}). The feature maps before patch embedding have large spatial scale and can result in huge computational overhead. For instance, LV-ViT \cite{jiang2021all} uses a 4-layer convolutional stem with 1.3G MAdds, which is too heavy regarding to our efficient inference objective. We accordingly proposed an efficient stem by reducing the channel dimension and replacing the intermediate convolutions using Depthwise Separable blocks \cite{howard2017mobilenets} with Depthwise and Pointwise convolution layers.

\textit{Dynamic Slicing in Patch Embedding.}
In vision transformer, an input image $\bm{x} \in \mathbb{R}^{H\times W\times 3}$ is first split into $N$ patches with size M, i.e. $\bm{x} \in \mathbb{R}^{N\times M\times 3}$. Then, these patches are projected to dimension $D$ by patch embedding layer with weight $\textbf{W}_{embed}\in \mathbb{R}^{3M\times D_{e}}$. To form a dynamic slice-able vision transformer supernet with variable embedding dimension $d_{e}$ ($d_{e} < D_{e}$), we make $\textbf{W}_{embed}$ slice-able in its output dimension:
\begin{equation}
\small
    \bm\psi_{embed}(\bm{x}) =  \textbf{W}_{embed}\slice{:d_{e}^{\bm\phi}} * \bm{x},
\end{equation}

\textit{Dynamic Slicing in Multi-Head Self Attention.}
Let $\bm{\mathcal{Q}} \in \mathbb{R}^{N\times d_h}$, $\bm{\mathcal{K}} \in \mathbb{R}^{N\times d_h}$ and $\bm{\mathcal{V}} \in \mathbb{R}^{N\times d_h}$ denote query, keys and values computed from the same sequence of inputs $\bm{\mathcal{X}} \in \mathbb{R}^{N\times d_{e}}$ with elastic embedding dimension $d_{e}$, using three slimmable linear transformations with weight $\textbf{W}_{q},\textbf{W}_{k},\textbf{W}_{v} \in \mathbb{R}^{D_{e}\times D_h}$, which are dynamic slice-able in both the input and output dimensions:
\begin{equation}
\left\{
\begin{aligned}
    &\bm{\mathcal{Q}} =  \textbf{W}_{q}\slice{:d_h^{\bm\phi}, :d} * \bm{\mathcal{X}},\\ 
    &\bm{\mathcal{K}} =  \textbf{W}_{k}\slice{:d_h^{\bm\phi}, :d} * \bm{\mathcal{X}},\\
    &\bm{\mathcal{V}} =  \textbf{W}_{v}\slice{:d_h^{\bm\phi}, :d} * \bm{\mathcal{X}}.\\
\end{aligned}
\right.
\end{equation}
Then, with the elastic tensors $\bm{\mathcal{Q}}, \bm{\mathcal{K}}, \bm{\mathcal{V}} \in \mathbb{R}^{N\times d_h}$, the slimmable self attention can be formulated as:
\begin{equation}
    \mathcal{A}_h = \mathtt{Softmax}(\bm{\mathcal{Q}} \bm{\mathcal{K}}^\top/\sqrt{d_h}) \bm{\mathcal{V}}.
\end{equation}

In multi-head self attention, multiple outputs in shape $N\times d_h$ are calculated in parallel via $h$ heads and then concatenated in shape of $N\times hd_h$ before projected back to $N\times d_{e}$. We, therefore, use a slimmable linear transformation $\textbf{W}_{proj}\in \mathbb{R}^{hD_h\times D_{e}}$ to project the $h$ elastic outputs from different heads back to the embedding dimension:
\begin{equation}
    \bm\psi_{attn}(\bm{\mathcal{X}}) =  \textbf{W}_{proj}\slice{:d_{e}, :hD_h} * \text{Concat}(\mathcal{A}_0,\dots, \mathcal{A}_{|h|}).
\end{equation}

\textit{Dynamic Slicing in MLP.}
In MLP blocks of vision transformers, we also apply the dynamic slicing scheme. The MLP block consists of two linear layers with an activation function $\sigma$, \textit{e.g.} GELU. Here, we use two slimmable linear layers to achieve dynamic MLP ratio:
\begin{equation}
\label{eqn:slimmableMLP}
\begin{aligned}
    &\bm\psi_{mlp}(\bm{\mathcal{X}}) =  \textbf{W}_{2}\slice{:d_{e}, :d_{mlp}^{\bm\phi}} * \sigma(\textbf{W}_{1}\slice{:d_{mlp}^{\bm\phi}, :d_{e}} * \bm{\mathcal{X}}).\\
\end{aligned}
\end{equation}

\subsection{Double-Headed Dynamic Gate}\label{sec:gate}
In this section, we present a dynamic gate with \textit{double-headed design}, serving as the routing agent $\Phi$ in Eq. \eqref{eqn:dyn_net} that predicts the states $\bm\phi$ for the sub-networks.

We formulate the dynamic routing as a categorical regression problem, and approximate the probability distribution function $p_{\bm\theta}(\cdot|\bm{\mathcal{X}})$.
To reduce the input feature map $\bm{\mathcal{X}}$ to a $|t|$-dimensional probability score vector for each route, we divide $p_{\bm\theta}(\cdot|\bm{\mathcal{X}})$ to two functions:
\begin{equation}
\small
    p_{\bm\theta}(\cdot|\bm{\mathcal{X}}) = \mathcal{F}(\mathcal{E}(\bm{\mathcal{X}})),
\end{equation}
where $\mathcal{E}$ is an encoder that reduces feature maps to a vector and the function $\mathcal{F}$ maps the reduced feature to a one-hot vector used for the subsequent channel slicing. Considering one of the routing gate in Fig. \ref{fig:dscnn}, given an input feature $\bm{\mathcal{X}}$ with dimension \begin{small}$c_{\mathcal{I}} \times H \times W$\end{small}, $\mathcal{E}(\bm{\mathcal{X}})$ reduces it to a vector \begin{small}$\bm{\mathcal{X}}_\mathcal{E} \in \mathbb{R}^{c_{\mathcal{I}}}$\end{small} which can be further mapped to a one-hot vector. 

Similar to prior works \cite{hu2019squeeze, yang2019gated} on channel attention and gating, we simply utilize average pooling as a light-weight encoder $\mathcal{E}$ to integrate spatial information:
\begin{equation}
\bm{\mathcal{X}}_{\mathcal{E}} = \frac{1}{H \times W}\sum_{i=1}^{H}\sum_{j=1}^{W}\bm{\mathcal{X}}\slice{:, i, j}
\end{equation}
As for feature mapping function $\mathcal{F}$, we adopt a two-layer dynamic MLP $\bm\psi_{mlp}$ with weights $\textbf{W}_1 \in \mathbb{R}^{C_{mlp} \times C_{\mathcal{I}}}$ and $\textbf{W}_2 \in \mathbb{R}^{|t| \times C_{mlp}}$ to predict scores for each of the $|t|$ routing choices. The probability of each routing choices is predicted by applying $\mathtt{softmax}$ function over the scores:

\begin{equation}
    \bm\phi \sim P_{\bm\theta}(\bm\phi = \bm e^{(t)}|\bm{\mathcal{X}}) = \frac{\text{exp}[\bm\psi^{(t)}_{mlp}(\bm\theta; \bm{\mathcal{X}}_{\mathcal{E}})]}{\bm\sum_t\text{exp}[\bm\psi^{(t)}_{mlp}(\bm\theta; \bm{\mathcal{X}}_{\mathcal{E}})]},
\end{equation}
where $\bm\phi$ is a \textit{one-hot} route indicator vector, and $\bm e^{(t)}$ is the $t$-th one-hot vector, denoting routing to the $t$-th path.

Our proposed channel gating function has a similar form with recent channel attention methods \cite{hu2019squeeze, yang2019gated}. The attention mechanism can be integrated into our gate with nearly zero cost, by adding another fully-connected layer with weights \begin{small}$\textbf{W}_3$\end{small} that projects the hidden vector with dimension $C_{mlp}$ back to the original channel number $C_{\mathcal{I}}$. Based on the conception above, we propose a \textit{double-headed dynamic gate} with a soft channel attention head and a hard routing head. The channel attention head can be defined as follows: 
\begin{equation}
    \widehat{\bm{\mathcal{X}}} = \bm{\mathcal{X}} \cdot \delta\big(\bm\psi_{mlp}(\textbf{W}_1, \textbf{W}_3; \bm{\mathcal{X}})\big),
\end{equation}
where \begin{small}$\delta (x) = 1 + \mathtt{tanh}(x)$\end{small} is the activation function adopted for the attention head.

\section{Training Dynamic Networks}\label{sec:methodology2}
\subsection{Disentangled Two-Stage Optimization Scheme}\label{sec:twostage}
We first formulate the optimization of dynamic network as follows:
\begin{equation}
\label{eqn:single_stage_optimization}
    \mathop{\min}_{\bm\theta}\mathop{\min}_\textbf{W}\mathbf{E}_{\bm\phi\sim p_{\bm\theta}(\cdot|\bm{x})}\big\{\mathcal{L}(\bm\phi, \textbf{W}; \bm{x})\big\},
\end{equation}
where $\mathcal{L}(\bm\phi, \textbf{W}; \bm{x})$ is the loss function of the dynamic network executing with the control signal $\bm\phi$. And we are interested in optimizing its expected value $\textbf{L}(\bm\theta, \textbf{W}; \bm{x}) := \mathbf{E}_{\bm\phi\sim p_{\bm\theta}(\cdot|\bm{x})}\big\{\mathcal{L}(\bm\phi, \textbf{W}; \bm{x})\big\}$.

\textit{Drawback of Single-Stage Optimization.} In previous dynamic inference methods \cite{hua2019channel,gao2018dynamic,Chen2019YouLT}, parameters of the routing agent $\bm\theta$ and parameters of the supernet $\textbf{W}$ are jointly optimized in an end-to-end fashion by directly optimizing Eq. \eqref{eqn:single_stage_optimization}. More specifically, these methods compute the gradient of $\textbf{W}$ and $\bm\theta$ by gradient estimators or reparameterization tricks. For instance, by using REINFORCE \cite{williams1992simple} and Monte Carlo sampling, gradient of $\textbf{W}$ and $\bm\theta$ can be estimated as follows:
\begin{equation}
\label{eqn:single_stage_gradient}
\small
\left\{
\begin{aligned}
    &\nabla_{\textbf{W}} \textbf{L}(\bm\theta, \textbf{W}; \bm{x}) \simeq \frac{1}{|k|} \sum_{k=1}^{|k|} \nabla_{\textbf{W}} \mathcal{L}(\bm\phi^k, \textbf{W}; \bm{x}),\\
    &\nabla_{\bm\theta} \textbf{L}(\bm\theta, \textbf{W}; \bm{x}) \simeq \frac{1}{|k|} \sum_{k=1}^{|k|}\mathcal{L}(\bm\phi^k, \textbf{W}; \bm{x}) \nabla_{\bm\theta}\log\big(p_{\bm\theta}(\bm\phi^k|\bm{x})\big),\\
\end{aligned}
\right.
\end{equation}
where $\bm\phi^k \sim p_{\bm\theta}(\cdot|\bm{x})$, i.d.d.

However, single stage training can cause generalization degradation and routing unfairness. First of all, as the routing signal is sampled from a conditional distribution of input $\bm{x}$, the input distribution for each route during training is also conditioned on the routing signal, \textit{i.e.} $\bm{x}_{\bm\phi}\sim p'_{\bm\theta}({\cdot|\bm\phi})$, which is to say, each route $\bm\psi$ in the dynamic supernet is activated for inputs that are not uniformly sampled from the data distribution. Similar issue has also been identified in \cite{han2021dynamic}. In an ideal case, easy samples tend to activate smaller sub-networks, while hard samples tend to use larger sub-networks. However, when the routing agent is not yet optimized, the input distribution for each route can be severely shifted from the data distribution and cause generalization degradation problems. Secondly, as the route in Eq. \eqref{eqn:single_stage_gradient} is activated by signal $\bm\phi^k \sim p_{\bm\theta}(\cdot|\bm{x})$, some route is activated for more training steps and can quickly gain advantage over other routes in early training stage, which further results in routing unfairness. Some of the previous works \cite{wu2018blockdrop,wang2020glance} on dynamic inference avoid these issues by pretraining $\textbf{W}$ before training $\bm\theta$ and $\textbf{W}$ jointly. Here, we adopt a two-stage optimization scheme to completely disentangle the optimization of $\bm\theta$ and $\textbf{W}$.

\textit{Disentangled Two-Stage Optimization.} 
To ensure the generality of every path and routing fairness, when optimizing $\textbf{W}$, the distribution of $\bm\phi$ should be independent to the input $\bm{x}$, \textit{e.g.} a uniform distribution $U(\bm{e})$ of one-hot vector $\bm{e}$. After the optimization of $\textbf{W}$, the routing gate can be trained with $\textbf{W}$ fixed as optimal. Based on the analysis above, we propose a two-stage optimization strategy in analogy to one-shot NAS \cite{brock2017smash,bender2018understanding}, to ensure the generality of every path in our DS-Net. The optimization of dynamic network (Eq. \eqref{eqn:single_stage_optimization}) is disentangled as follows:
\begin{equation}
\left\{
\begin{aligned}
    &\textbf{W}^{\star} = \mathop{\arg\min}^{~}_\textbf{W}\mathbf{E}_{\bm\phi\sim U(\bm e)}\big\{\mathcal{L}(\bm\phi, \textbf{W}; \bm{x})\big\}\\
    &\bm\theta^{\star} = \mathop{\arg\min}_{\bm\theta}\mathbf{E}_{\bm\phi\sim p_{\bm\theta}(\cdot|\bm{x})}\big\{\mathcal{L}(\bm\phi, \textbf{W}^{\star}; \bm{x})\big\}\\
\end{aligned}
\right.
\end{equation}
Here, the optimization of supernet parameters $\textbf{W}$ is completely isolated from the dynamic gate $\Phi(\bm\theta, \bm{x})$, in contrast to Eq. \eqref{eqn:single_stage_optimization} where these two are highly entangled. Practically, in \textbf{Stage I}, we disable the routing gate and train a weight sharing supernet to reach the optimal $\textbf{W}^{\star}$ by sampling sub-networks; then in \textbf{Stage II}, we fix the optimal parameters $\textbf{W}^{\star}$ of the supernet and optimize the parameters $\bm\theta$ of the dynamic routing gates.
\begin{figure}
    \centering
    \includegraphics[width=1.0\linewidth]{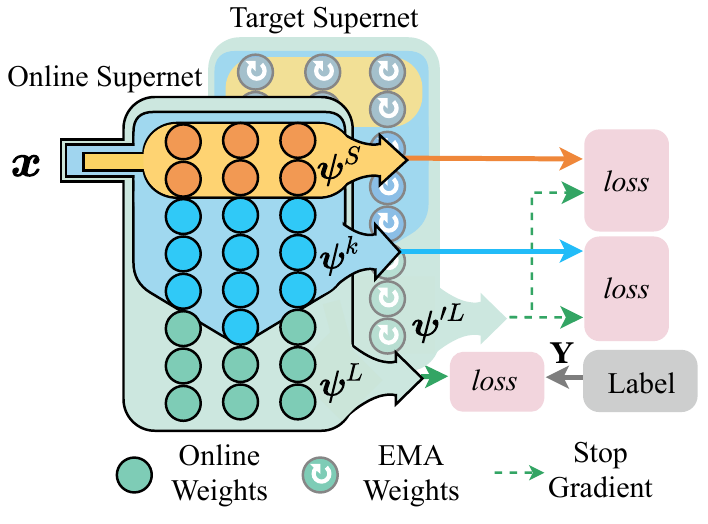}
    \caption{Training process of dynamic supernet with in-place bootstrapping.}\label{fig:ib_scheme}
\end{figure}

\subsection{Training Techniques for Anytime Supernet}
The \emph{sandwich rule} and \emph{in-place distillation (ID)} techniques \cite{Yu2019UniversallySN} proposed for Universally Slimmable Networks enhanced their overall performance. Following \emph{sandwich rule}, in each training step, the smallest sub-network, largest sub-network and $|k|$ random sub-networks are sampled for training. The expected loss of the dynamic supernet is estimated as follows:
\begin{equation}
    \mathbf{E}_{\bm\phi}\big\{\mathcal{L}\big\} \simeq \frac{1}{|k|+2} \left(\mathcal{L}^S + \mathcal{L}^L + \sum_{k=1}^{|k|}\mathcal{L}^k\right),
\end{equation}
where $\mathcal{L}^S$, $\mathcal{L}^L$ and $\mathcal{L}^k$ are loss function of the smallest sub-network, largest sub-network and $|k|$ uniformly sampled sub-networks, respectively.

In \textit{in-place distillation}, the largest sub-network is used as the target network generating soft labels for other sub-networks. Let $\mathcal{Y}$ denote the output of the network and $\bm{y}$ denote the true labels, the in-place distillation losses can be summarized as follows:
\begin{equation}
\left\{
\begin{aligned}
    &\mathcal{L}_\textit{ID}^{L} = \mathcal{L}_\textit{CE}(\bm\psi^{L}(\bm{x}), \bm{y}),\\ 
    &\mathcal{L}_\textit{ID}^{k} = \mathcal{L}_\textit{CE}(\bm\psi^k(\bm{x}), \bm\psi^{L}(\bm{x})),\\
    &\mathcal{L}_\textit{ID}^{S} = \mathcal{L}_\textit{CE}(\bm\psi^{S}(\bm{x}), \bm\psi^{L}(\bm{x})),\\
\end{aligned}
\right.
\end{equation}
where \begin{small}$\mathcal{L}_\textit{CE}(\bm{p}, \bm{q}) = -\sum_i \bm{q}_i\log(\bm{p}_i)$\end{small} is the cross-entropy loss of output $\bm{p}$ and target $\bm{q}$, where we always stop the gradient computation of the target $\bm{q}$.

However, as observed in BigNAS \cite{Yu2020BigNASSU}, training a more complex model with in-place distillation could be highly unstable. Without residual connection and special weight initialization tricks, the loss exploded at the early stage and can never converge. Acute fluctuation appeared in the weight of the largest sub-network can cause convergence hardship, especially in the early stage of training. To overcome the convergence hardship in slimmable networks and improve the overall performance of our supernet, we propose a series of training techniques, including \textit{in-place bootstrapping (IB)} with its two variations and \textit{multi-view consistency (MvCo)} in two forms.

\subsubsection{In-place Bootstrapping}
\textit{In-place Bootstrapping (IB).}
In recent years, a growing number of self-supervised methods with bootstrapping \cite{BYOL, guo2020bootstrap, deepcluster} and semi-supervised methods based on consistency regularization \cite{laine2016temporal,tarvainen2017mean} use historical representations to produce targets for the online network. Inspired by this, we propose to bootstrap on previous representations in our \textit{supervised} in-place distillation training. As shown in Fig. \ref{fig:ib_scheme}, we use the exponential moving average (EMA) of the supernet as the target supernet that generates soft labels. Let $\textbf{W}$ and $\textbf{W}'$ denote the parameters of the online supernet and the target supernet, respectively. We have:
\begin{equation}
\begin{aligned}
    \textbf{W}_{\tau}' = \alpha\textbf{W}_{{\tau}-1}' + (1-\alpha)\textbf{W}_{\tau},
\end{aligned}
\end{equation}
where $\alpha$ is a momentum factor controlling the ratio of the historical parameter and ${\tau}$ is a training timestamp which is usually measured by a training iteration. During training, the EMA supernet is more robust to the acute fluctuation in the parameter of online supernet, thus can consistently provide high quality target for the smaller sub-networks. Let $\bm\psi'$ denote the routes in the target supernet with EMA weight $\textbf{W}'$, the training losses of in-place bootstrapping can be denoted as follows:
\begin{equation}
\label{eqn:EB}
\left\{
\begin{aligned}
    &\mathcal{L}_\textit{IB}^{L} = \mathcal{L}_\textit{CE}(\bm\psi^{L}(\bm{x}), \bm{y}),\\ 
    &\mathcal{L}_\textit{IB}^k = \mathcal{L}_\textit{CE}(\bm\psi^k(\bm{x}), \bm\psi'^{L}(\bm{x})),\\
    &\mathcal{L}_\textit{IB}^{S} = \mathcal{L}_\textit{CE}(\bm\psi^{S}(\bm{x}), \bm\psi'^{L}(\bm{x})).\\
\end{aligned}
\right.
\end{equation}

\textit{Ensemble IB.}
As pointed out in \cite{Meal, Mealv2}, an ensemble of teacher networks can generate more diverse, more accurate and more general soft labels for distillation training of the student network. In our supernet, there are tons of sub-models with different architectures, which can generate different soft labels. To take advantages of this, we use different sub-networks as a teacher ensemble when performing in-place bootstrapping for the smallest sub-network, forming the ensemble in-place bootstrapping (E-IB) scheme. More specifically, the smallest sub-network is trained to predict the probability ensemble of all the other sampled sub-networks in the target network:
\begin{equation}
\begin{aligned}
    \widehat{\bm\psi'}(\bm{x}) = \frac{1}{|k|+1}\left(\bm\psi'^{L}(\bm{x})+ \sum_{k=1}^{|k|}\bm\psi'^k(\bm{x})\right).
\end{aligned}
\end{equation}
To sum up, the E-IB scheme replace the $\mathcal{L}_\textit{IB}^{S}$ in Eq. \eqref{eqn:EB} with $\mathcal{L}_\textit{E-IB}^{S}$, which is calculated by:
\begin{equation}
    \mathcal{L}_\textit{E-IB}^{S} = \mathcal{L}_\textit{CE}(\bm\psi^{S}(\bm{x}), \widehat{\bm\psi'}(\bm{x})).\\
\end{equation}

\textit{Hierarchical IB with Slimmable Projector.}
As we expand the dynamic routing space to more dimensions, \textit{e.g.} depth, kernel size, \textit{etc.}, the convergence hardship becomes more severe. The small sub-networks converge much slower than the larger ones. To reduce the gap between sub-networks and ease the convergence hardship, we add intermediate supervision after each stage of the supernet, forming the hierarchical in-place bootstrapping (H-IB). 
To relax the strict per-pixel matching constraint and tackle the size mismatch issue of intermediate output of different sub-networks, we introduce a slimmable predictor $\bm\rho$ for online supernet. Consider $\ell$-th intermediate features after global average pooling: $\bm\psi^k_\ell(\bm{x})\in \mathbb{R}^{c}$ and $\bm\psi'^L_\ell(\bm{x})\in \mathbb{R}^{C}$ from the $k$-th sampled sub-network of online supernet and the largest sub-network of target supernet, respectively. We use a slimmable MLP (defined as Eq. \eqref{eqn:slimmableMLP}) with weight $\textbf{W}_1\in \mathbb{R}^{C_{mlp}\times C}$ and $\textbf{W}_2\in \mathbb{R}^{C\times C_{mlp}}$ as the predictor $\bm\rho$ to predict the target $\bm\psi'^L_\ell(\bm{x})$ from the online feature $\bm\psi^k_\ell(\bm{x})$.
Similar to previous work on intermediate distillation \cite{Romero2014FitNetsHF}, we use the simplistic L2 distance as the loss function, the H-IB loss of the $k$-th sampled sub-network is formulated as follows:
\begin{equation}\label{eqn:h_ib_1}
\begin{aligned}
    \mathcal{L}^k_\textit{H-IB} = \frac{1}{|\ell|}\sum^{|\ell|}_{\ell=1}\frac{1}{C} \bigg\| \bm\rho\big(\bm\psi^k_\ell(\bm{x})\big) - \bm\psi'^L_\ell(\bm{x}) \bigg\|_2^2.
\end{aligned}
\end{equation}
The H-IB loss for the smallest sub-network $\bm\psi^S$ has the same form. When using H-IB scheme, we always balance the loss with the end-to-end bootstrapping loss $\mathcal{L}_\textit{IB}$ or $\mathcal{L}_\textit{E-IB}$. Taking original IB as an example, the total loss of H-IB scheme for the $k$-th sampled sub-network of online supernet is:
\begin{equation}\label{eqn:h_ib_2}
    \mathcal{L} = \lambda_\textit{IB} \mathcal{L}_\textit{IB} + \lambda_\textit{H-IB} \mathcal{L}_\textit{H-IB}
\end{equation}

\subsubsection{Multi-view Consistency}
In sandwich rule, the output $\bm\psi(\bm x)$ of each sub-networks are calculated using the same input $\bm x$. As the consistency over different view of the same image is important for the generality of the network, we take the advantage of multi-path sampling in sandwich rule, and use different views of a training sample for each sampled paths.

\textit{In-place Bootstrapping with Multi-view Consistency.}
In in-place bootstrapping scheme, given a training sample $\bm x$, multi-view consistency (MvCo) is performed by sampling $|k|+2$ augmented views \begin{small}$\{\bm x_k\}, \bm{x}_S, \bm{x}_L \sim \boldsymbol p_{aug}(\cdot|\bm x)$\end{small} for the $|k|+2$ sub-networks. Overall, in-place bootstrapping with multi-view consistency is formulated as:
\begin{equation}
\left\{
\begin{aligned}
    &\mathcal{L}_\textit{IB}^{L} = \mathcal{L}_\textit{CE}\bigl(\bm\psi^{L}(\bm x_L), \bm{y}\bigr),\\ 
    &\mathcal{L}_\textit{IB}^k = \mathcal{L}_\textit{CE}\bigl(\bm\psi^k(\bm x_k), \bm\psi'^{L}(\bm x_L)\bigr),\\
    &\mathcal{L}_\textit{IB}^{S} = \mathcal{L}_\textit{CE}\bigl(\bm\psi^{S}(\bm x_S), \bm\psi'^{L}(\bm x_L)\bigr).\\
\end{aligned}
\right.
\end{equation}

\textit{External Distillation with Multi-view Consistency.}
As observed in \cite{touvron2020deit}, distillation schemes designed for CNNs are not optimal for vision transformers. Similarly, in-place distillation and in-place bootstrapping designed for slimmable CNNs is not suitable for dynamic slimmable transformer. The training can not even converge if naively apply in-place distillation on transformers (see Section \ref{sec:exp_vit_training}).
During in-place distillation training, the cross-entropy between smaller sub-networks and the largest one becomes dominant before the largest network learning any useful knowledge.
This can be attributed to the slower convergence of vision transformers, in contrast to CNNs with inductive bias. An intuitive solution is to pretrain the largest model before applying in-place distillation. As an alternative, we directly adopt a well-trained teacher model $\mathcal{T}$ as an external supervision. 

Similar to the case of in-place bootstrapping, to promote multi-view consistency, we sampling $|k|+3$ augmented views \begin{small}$\{\bm x_k\}, \bm{x}_S, \bm{x}_L, \bm{x}_T \sim \boldsymbol p_{aug}(\cdot|\bm x)$\end{small} for the $|k|+2$ sub-networks and the teacher model. We use a similar distillation scheme with DeiT \cite{touvron2020deit}. Each of the sampled paths learns to predict the hard label generated by the teacher $\bm{y}_\mathcal{T} = \mathtt{argmax}\big(\mathcal{T}(\bm x_T)\big)$, and ground truth label $\bm{y}$. Losses of external distillation with multi-view consistency for dynamic transformers are as follows:
\begin{equation}
\left\{
\begin{aligned}
    &\mathcal{L}_\textit{ED}^{L} = \frac{1}{2}\Big(\mathcal{L}_\textit{CE}\bigl(\bm\psi^{L}(\bm x_L), \bm{y}\bigr) + \mathcal{L}_\textit{CE}\bigl(\bm\psi^{L}(\bm x_L), \bm{y}_\mathcal{T}\bigr)\Big),\\ 
    &\mathcal{L}_\textit{ED}^k = \frac{1}{2}\Big(\mathcal{L}_\textit{CE}\bigl(\bm\psi^{k}(\bm x_k), \bm{y}\bigr) + \mathcal{L}_\textit{CE}\bigl(\bm\psi^k(\bm x_k), \bm{y}_\mathcal{T}\bigr)\Big),\\
    &\mathcal{L}_\textit{ED}^{S} = \frac{1}{2}\Big(\mathcal{L}_\textit{CE}\bigl(\bm\psi^{S}(\bm x_S), \bm{y}\bigr) + \mathcal{L}_\textit{CE}\bigl(\bm\psi^{S}(\bm x_S), \bm{y}_\mathcal{T}\bigr)\Big).\\
\end{aligned}
\right.
\end{equation}

\subsection{Training Techniques for Input-Dependent Gate}

In training stage II, we propose to use the end-to-end classification cross-entropy loss $\mathcal{L}_\textit{CLS}$ and a complexity penalty loss $\mathcal{L}_\textit{CP}$ to train the gate, aiming to choose the most efficient and effective sub-networks for each instance.

\textit{Differentiable Relaxation with Gumbel-Softmax.}
To optimize the non-differentiable routing head of dynamic gate in an end-to-end fashion with classification loss $\mathcal{L}_\textit{CLS}$, we use $\mathtt{gumbel}$-$\mathtt{softmax}$ \cite{jang2016categorical,maddison2017concrete}, a classical way to optimize neural networks with $\mathtt{argmax}$ or with discrete variables sampled from $\mathtt{softmax}$, \textit{e.g.} the one-hot vector $\bm\phi$ in this work, by relaxing $\mathtt{argmax}$ or discrete sampling to differentiable $\mathtt{softmax}$ in gradient computation. 

\textit{Complexity penalty.} Complexity penalty loss $\mathcal{L}_\textit{CP}$ is used to increase the model efficiency in training stage II. To provide a stable and fair constraint, we use the number of multiply-adds on the fly, $\mathtt{MAdds}(\bm\theta;\bm{x})$, as the metrics of model complexity. Specifically, the complexity penalty is given by:
\begin{equation}
    \mathcal{L}_\textit{CP}= \big(\frac{\mathtt{MAdds}(\bm\theta;\bm{x})}{\mathbf{T}}\big)^2,
\end{equation}
where $\mathbf{T}$ is a normalize factor set to the total MAdds of the supernet in our implementation. Note that this loss term always pushes the gate to route towards a faster architecture, rather than towards an architecture with target MAdds, which can effectively prevent routing easy and hard instances to the same architecture.

\textit{sandwich gate sparsification.}
When only using the classification loss, we empirically found that the gate easily collapses into a static one even if we add Gumbel noise \cite{jang2016categorical} to help the optimization of $\mathtt{gumbel}$-$\mathtt{softmax}$. Apparently, using only $\mathtt{gumbel}$-$\mathtt{softmax}$ technique is not enough for this multi-objective dynamic gate training. 
To further overcome the convergence hardship and increase the dynamic diversity of the gate, a technique named \emph{sandwich gate sparsification (SGS)} is further proposed. We use the smallest sub-network and the whole network to identify easy and hard samples online and further generate the ground truth routing signals for the routing heads of all the dynamic gates.

Guaranteed by generalized residual learning Eq. \eqref{eqn:residual}, larger sub-networks should always be more accurate because the accuracy of smaller ones can always be achieved by learning new connections to zeros. Thus, given a well-trained supernet, input samples can be roughly classified into three difficulty levels: \textbf{a) Easy samples $\bm{x}_\textit{easy}$} that can be correctly classified by the smallest sub-network; \textbf{b) Hard samples $\bm{x}_\textit{hard}$} that can not be correctly classified by the largest sub-network; \textbf{c) Dependent samples $\bm{x}_\textit{dep}$}: Other samples in between. In order to minimize the computation cost, easy samples should always be routed to the smallest sub-network (\textit{i.e.} gate target \begin{small}$\bm{y}_{\Phi}(\bm{x}_\textit{easy}) = \bm{e}^1$\end{small}). For dependent samples and hard samples, we always encourage them to pass through the largest sub-network, even if the hard samples can not be correctly classified (\textit{i.e.} \begin{small}$\bm{y}_{\Phi}(\bm{x}_\textit{hard}) = \bm{y}_{\Phi}(\bm{x}_\textit{dep}) = \bm{e}^{|t|}$\end{small}). Another gate target strategy is also discussed in Section \ref{sec:SGS_strategy}.

Based on the generated gate target, we define the SGS loss as the cross-entropy between the predicted probability for each choice $p_{\bm\theta}(\cdot|\bm{x})$ and the generated ground truth choice $\bm{y}_{\Phi}(\bm{x})$:
\begin{equation}
\mathcal{L}_\textit{SGS} = \mathcal{L}_\textit{CE}\big(p_{\bm\theta}(\cdot|\bm{x}), \bm{y}_{\Phi}(\bm{x})\big).
\end{equation}

Overall, the routing gate can be optimized with a joint loss function:
\begin{equation}
    \mathcal{L} = \lambda_1 \mathcal{L}_\textit{CLS} + \lambda_2 \mathcal{L}_\textit{CP} + \lambda_3 \mathcal{L}_\textit{SGS}.
\end{equation}

\section{Experiments}\label{sec:experiments}
\begin{table}[t]
\centering
\caption{Routing space of DS-MBNet and DS-MBNet++. (k: kernel size, c: output channel number, n: layer number, s: stride, p: padding size)}\label{tab:dsmb_space}
\centering
\footnotesize
\setlength{\tabcolsep}{8pt}
\begin{tabular}{llllll}
\toprule
block       & k         & c                 & n         & s  & p\\
\midrule
Conv        & 3         & 16                & 1         & 2  & 1\\
DSConv 	    & 3 	    & 32                & 1         & 1  & 1\\
DSConv	    & 3 	    & 48                & 2         & 2  & 1\\
DSConv	    & 3	        & 96                & 2         & 2  & 1\\
DSConv	    & 3 	    & [224 : 640 : 32]  & 6         & 2  & 1\\
DSConv	    & 3 	    & [736 : 1152 : 32] & 2         & 2  & 1\\
Pool + FC   & -         & 1000              & 1         & -  & -\\
\midrule
Conv        & 3         & 16                & 1         & 2  & 1\\
DSConv 	    & 5 	    & 32                & 1         & 1  & 1\\
DSConv	    & 5 	    & 48                & 2         & 2  & 1\\
DSConv	    & 5	        & 96                & 2         & 2  & 1\\
DSConv	    & \{3, 5, 7\} & [288 : 640 : 32]  & 6       & 2  & [0 : 3]\\
DSConv	    & \{3, 5, 7\} & [800 : 1152 : 32] & 2       & 2  & [0 : 3]\\
Pool + FC   & -         & 1000              & 1         & -  & -\\
\bottomrule
\end{tabular}
\end{table}
\begin{table}[t]

\centering
\caption{Routing space of DS-ResNet and DS-ResNet++. (k: kernel size, c: output channel number, n: layer number, s: stride, p: padding size)}\label{tab:dsres_space}
\centering
\footnotesize
\setlength{\tabcolsep}{9.5pt}
\begin{tabular}{llllll}
\toprule
block       & k         & c                 & n         & s  & p\\
\midrule
Conv        & 3         & 64                & 1         & 2  & 1\\
MaxPool     & 3         & 64                & 1         & 2  & 1\\
ResBlock 	& 3 	    & [64 : 256 : 64]   & 3         & 1  & 1\\
ResBlock	& 3 	    & [128 : 512 : 128] & 4         & 2  & 1\\
ResBlock	& 3	        & [256 : 1024 : 256]& 6         & 2  & 1\\
ResBlock	& 3 	    & [512 : 2048 : 512]& 3         & 2  & 1\\
Pool + FC   & -         & 1000              & 1         & -  & -\\
\midrule
Conv        & 3         & 64                & 1         & 2  & 1\\
MaxPool     & 3         & 64                & 1         & 2  & 1\\
ResBlock 	& 3 	    & [128 : 256 : 32]  & [2 : 3]   & 1  & 1\\
ResBlock	& 3 	    & [256 : 512 : 32]  & [2 : 4]   & 2  & 1\\
ResBlock	& 3	        & [512 : 1024 : 64] & [5 : 8]   & 2  & 1\\
ResBlock	& 3 	    &[1024 : 2048 : 128]& [2 : 3]   & 2  & 1\\
Pool + FC   & -         & 1000              & 1         & -  & -\\
\bottomrule
\end{tabular}
\end{table}
\begin{table}[t]

    \centering
    \caption{Routing space of DS-ViT++.}
    \label{tab:space_vit}
    \setlength{\tabcolsep}{20pt}
    \begin{tabular}{l|c}
        \toprule
        Embed Dim ($d_{e}$) & [192 : 32 : 416] \\
        Q-K-V Dim ($d_{a}$) & [192 : 48 : 384] \\
        Head Num ($h$)     & [4 : 1 : 8] \\
        MLP Ratio ($d_\textit{MLP}/d_{e}$) & [3 : 0.5 : 4] \\
        Depth ($l$)    & [12 : 1 : 14] \\
        \bottomrule    
    \end{tabular}
\end{table}

\subsection{Dataset.} We evaluate our method on a large scale classification dataset, ImageNet \cite{deng2009imagenet}, two transfer learning datasets, CIFAR-10 and CIFAR-100 \cite{Krizhevsky09cifar} and an object detection dataset, PASCAL VOC \cite{everingham2010pascal}. ImageNet contains 1.2 M $\mathtt{train}$ set images and 50 K $\mathtt{val}$ set images in 1000 classes. We use all the training data in both of the two training stages. Our results are obtained on the $\mathtt{val}$ set with image size of $224\times224$. We test the transferability of our DS-Net on CIFAR-10 and CIFAR-100, which be comprised of 50,000 training and 10,000 test images, in 10 classes and 100 classes, respectively. Note that few previous works on dynamic networks and network pruning reported results on object detection. We take PASCAL VOC, one of the standard datasets for evaluating object detection performance, as an example to further test the generality of our dynamic networks on object detection. All the detection models are trained with the combination of 2007 $\mathtt{trainval}$ and 2012 $\mathtt{trainval}$ set and tested on VOC 2007 $\mathtt{test}$ set.

\subsection{Architecture details.} Following previous works on static and dynamic network pruning, we use two representative CNN networks to evaluate our method, \textit{i.e.}, the lightweight non-residual network MobileNetV1 \cite{howard2017mobilenets} and the residual network ResNet 50 \cite{he2016deep}. We further evaluate our method on a representative vision transformer model, ViT \cite{dosovitskiy2021vit}.
\subsubsection{DS-Net}
We first present our basic dynamic slimmable CNN models with dynamic filter size.

\textit{DS-MBNet.} In Dynamic Slimmable MobileNetV1 (DS-MBNet), we insert our double-headed gate in each residual blocks and only enable one routing gate after the fifth depthwise separable convolution block. Specifically, a fixed filter number is used in the first 5 blocks while the width of the rest 8 blocks are controlled by the gate. The total routing space contains 14 paths with different filter number.

\textit{DS-ResNet.} In Dynamic Slimmable ResNet 50 (DS-ResNet), we enable the routing gate in the first block of each stage. Each one of those blocks contains a skip connection with a projection layer, \textit{i.e.} $1\times1$ convolution. The filter number of this projection convolution is also controlled by the gate to avoid channel inconsistency when adding skip features with residual output. In other residual blocks, the routing heads of the gates are disabled and all the layers in those blocks inherit the control signal $\bm\phi$ of the first blocks of each stage. To sum up, there are 4 gates (one for each stage) with both heads enabled. Every gates have 4 equispaced candidate slimming ratios, \textit{i.e.} $\{0.25, 0.5, 0.75, 1\}$. The total routing space contains $4^4 = 256$ possible paths with different computation complexities.

\subsubsection{DS-Net++}
We further present DS-Net++ models, where ``++'' represents adding the enlarged compound routing space and improved training techniques.

\textit{DS-MBNet++.} Similar to DS-MBNet, we use one routing gate after the fifth depthwise separable convolution block of DS-MBNet++. The gate controls the kernel size, channel number and padding size of the subsequent blocks. As dynamic depth is not suitable for non-residual networks following GUIDELINE I, we do not use dynamic depth in DS-MBNet++. The routing space contains 12 paths in total. Table \ref{tab:dsmb_space} summarizes the detailed routing space of DS-MBNet and DS-MBNet++.

\textit{DS-ResNet++.} We enable one routing gate in the first residual block of DS-ResNet++, which controls the channel number and layer number of the four stages. The routing space contains 9 paths in total. Table \ref{tab:dsres_space} summarizes the detailed routing space of DS-ResNet and DS-ResNet++.

\textit{DS-ViT++.} In Dynamic Slimmable Vision Transformer (DS-ViT++), we use one routing gate after efficient stem and before patch projection. The gate controls the embedding dimension, Q-K-V dimension, head number, MLP ratio and depth of the DS-ViT supernet. The routing space is shown in Table \ref{tab:space_vit}.
\begin{figure*}[t]
    \centering
    \vspace{-.8 em}
    \includegraphics[width=0.9\linewidth]{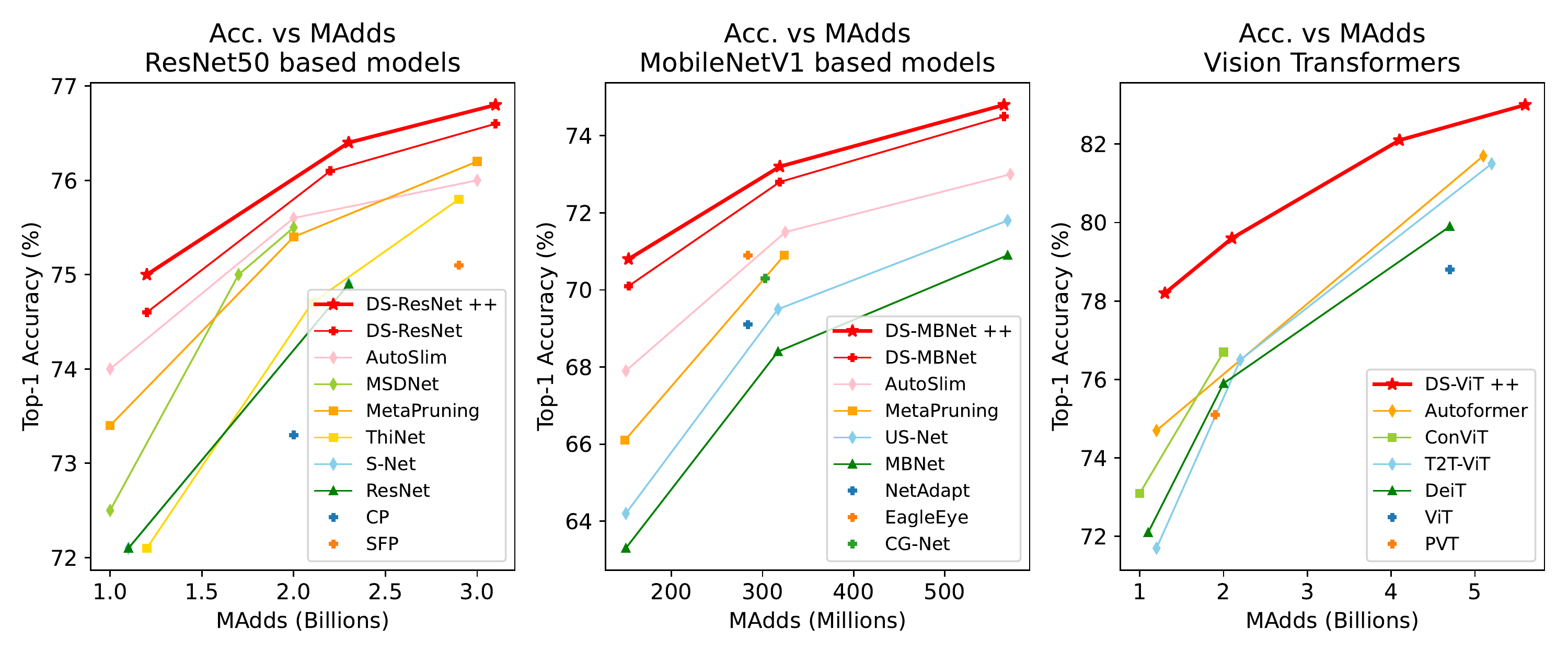}
    \vspace{-.8 em}
    \caption{Accuracy \textit{vs.} complexity on ImageNet.}
    \vspace{-.8 em}
    \label{fig:imagenet}
\end{figure*}

\subsection{Training details.}
We train our CNN supernets with 512 total batch size on ImageNet, using SGD optimizer with 0.08 initial learning rate for DS-MBNet and 0.2 initial learning rate for DS-ResNet, respectively. DS-MBNet and DS-ResNet are trained with \textit{E-IB} scheme, while DS-MBNet++ and DS-ResNet++ are trained with \textit{H-IB + MvCo}. The number of random sub-networks of sandwich sampling is set to $|k|=2$ for all the CNN supernets. The momentum factor $\alpha$ of EMA teacher is set to 0.997. We use cosine learning rate scheduler with 5 warm up epochs to adjust the learning rate in 150 epochs. Weight decay is set to 1$e$-4. Other settings are the same as previous works on slimmable networks \cite{Yu2019SlimmableNN,Yu2019UniversallySN,yu2019autoslim}. To stabilize the optimization, weight decay of all the layers in the dynamic gate is disabled. The weight $\bm\gamma$ of the last normalization layer of each residual block is initialized to zeros following \cite{Yu2020BigNASSU}. Additional training techniques include \cite{goyal2017accurate,cubuk2019autoaugment}.

The vision transformer supernet is trained with \textit{ED + MvCo} scheme using a similar recipe as DeiT \cite{touvron2020deit} with RegNetY-16GF \cite{radosavovic2020designing} as the teacher. The number of random sub-networks of sandwich sampling is set to $|k|=1$. We use AdamW optimizer with 1$e$-3 initial learning rate for a total batch size of 1,024. The weight decay is set to 5$e$-2. In 300 training epochs, the learning rate first linearly warms up for 5 epochs and then decays with a cosine scheduler. We use similar regularization and augmentation as DeiT \cite{touvron2020deit}, including label smoothing, stochastic depth, RandAugment, Cutmix, Mixup, random erasing. Different from them, we do not use repeated augmentation.

For gate training, we use SGD optimizer with 0.05 initial learning rate for a total batch size of 512. The learning rate decays to 0.9$\times$ of its value in every epoch. It takes 10 epochs for the gate to converge. The 3 loss balancing factors are set to \begin{small}$\lambda_1 = 1$, $\lambda_2 = 0.5$, $\lambda_3 = 1$\end{small} in our experiments. Different target MAdds is reached by adjusting the routing space during gate training. For instance, when training the gate of DS-MBNet-S, we only use the first 4 sub-networks.

For transfer learning experiments on CIFAR datasets, we follow similar settings with \cite{Kornblith2018DoBI} and \cite{Huang2019GPipeET}. We transfer our supernet for 70 epochs including 15 warm-up epochs and use cosine learning rate scheduler with an initial learning rate of 0.7 for a total batch size of 1024. 

For object detection task, we train all the networks following \cite{liu2016ssd} and \cite{li2017fssd} with a total batch size of 128 for 300 epochs. The learning rate is set to 0.004 at the first, then divided by 10 at epoch 200 and 250.

\subsection{Main Results on ImageNet}
\subsubsection{DS-CNNs and DS-CNNs++}

We validate the effectiveness of DS-Net and DS-Net++ with two CNN networks on ImageNet. The performance comparison of models based on MobileNet and ResNet are shown in Table \ref{tab:imagenet_mbnet} and Table \ref{tab:imagenet_resnet}, respectively. Fig. \ref{fig:imagenet} provides a more intuitive accuracy-efficiency comparison. DS-Net++ improves over our strong basic methods DS-Net by a clear margin (up to \textbf{0.7\%} Top-1 Accuracy) with both ResNet and MobileNet. DS-Net and DS-Net++ models with different computation complexity consistently outperforms recent static pruning methods, dynamic inference methods and NAS methods. We further make detailed comparison with these methods.

First, our DS-Net and DS-Net++ models achieve 2-4$\times$ computation reduction over MobileNetV1 (70.9\% \cite{howard2017mobilenets}) and ResNet-50 (76.1\% \cite{he2016deep}) with minimal accuracy drops. For example, DS-MBNet-S++ reduce the computation complexity of MobileNetV1 by \textbf{3.65$\times$} with only 0.1\% accuracy drop.  We also tested the real world latency on efficient networks. Compared to the ideal acceleration tested on channel scaled MobileNetV1, which is 1.91$\times$, our DS-MBNet and DS-MBNet++ achieves comparable 1.62$\times$ and 1.54$\times$ acceleration with much better performance.
In particular, our DS-MBNet++ surpasses the original and the channel scaled MobileNetV1 \cite{howard2017mobilenets} by \textbf{7.5\%}, \textbf{4.8\%} and \textbf{3.9\%} with similar MAdds and minor increase in latency.\footnote{The increase in latency is possibly caused by using a complexity penalty $\mathcal{L}_\textit{CP}$ based on MAdds. Here, we only demonstrate that our dynamic slicing scheme can achieve comparable acceleration with the ideal case and leave further optimization on latency as a future work.}
\begin{table}[t]
  \caption{Comparison of efficient inference methods with MobileNet \cite{howard2017mobilenets} on ImageNet. \textcolor{DarkOrange3}{Orange}: NAS or pruning methods, \textcolor{DeepSkyBlue4}{Blue}: dynamic inference methods,  \textcolor{purple}{Red}: our method.
  }
  \label{tab:imagenet_mbnet}
  \centering
  \begin{tabular}{c|l|c|c|c}
    \toprule
    \multicolumn{2}{c|}{Method}      & MAdds & Latency & Top-1 \\
    \toprule
    \multirow{5}{*}{\makecell{500M\\MAdds}}
    & MBNetV1 1.0$\times$ \cite{howard2017mobilenets}                                   & 569M & 63ms & 70.9 \\
    & US-MBNetV1~\pub{ICCV19}~\cite{Yu2019UniversallySN}                                 & 569M & \na & 71.8 \\
    & \textcolor{DarkOrange3}{AutoSlim}~\pub{arxiv19}~\cite{yu2019autoslim}                                & 572M & \na & 73.0 \\
    & \textcolor{purple}{DS-MBNet-L (Ours)}                                             & 565M & 69ms & 74.5\\
    & \textcolor{purple}{DS-MBNet-L++ (Ours)}                                             & 570M & 73ms & \textbf{74.8}\\
    \midrule
    \multirow{9}{*}{\makecell{300M\\MAdds}}
    & MBNetV1 0.75$\times$ \cite{howard2017mobilenets}                                  & 317M & 48ms & 68.4 \\
    & US-MBNetV1~\pub{ICCV19}~\cite{Yu2019UniversallySN}                                & 317M & \na & 69.5 \\
    & \textcolor{DarkOrange3}{NetAdapt}~\pub{ECCV18}~\cite{yang2018netadapt}                               & 284M & \na & 69.1 \\
    & \textcolor{DarkOrange3}{Meta-Pruning}~\pub{ICCV19}~\cite{Liu2019MetaPruningML}                        & 281M & \na & 70.6 \\
    & \textcolor{DarkOrange3}{EagleEye}~\pub{ECCV20}~\cite{li2020eagleeye}                                  & 284M & \na & 70.9 \\
    & \textcolor{DarkOrange3}{AutoSlim}~\pub{arxiv19}~\cite{yu2019autoslim}                                & 325M & \na & 71.5 \\
    & \textcolor{DeepSkyBlue4}{CG-Net-A}~\pub{NeurIPS19}~\cite{hua2019channel}                                 & 303M & \na & 70.3 \\
    & \textcolor{purple}{DS-MBNet-M (Ours)}                                             & 319M & 54ms & 72.8\\
    & \textcolor{purple}{DS-MBNet-M++ (Ours)}                                             & 325M & 57ms & \textbf{73.2}\\
    \midrule
    \multirow{5}{*}{\makecell{150M\\MAdds}}
    & MBNetV1 0.5$\times$ \cite{howard2017mobilenets}                                   & 150M & 33ms & 63.3 \\
    & US-MBNetV1~\pub{ICCV19}~\cite{Yu2019UniversallySN}                                 & 150M & \na & 64.2 \\
    & \textcolor{DarkOrange3}{AutoSlim}~\pub{arxiv19}~\cite{yu2019autoslim}                                & 150M & \na & 67.9 \\
    & \textcolor{purple}{DS-MBNet-S (Ours)}                                             & 153M & 39ms & 70.1\\
    & \textcolor{purple}{DS-MBNet-S++ (Ours)}                                             & 156M & 41ms & \textbf{70.8}\\
    
    \bottomrule
  \end{tabular}
\end{table}
Second, DS-Net and DS-Net++ consistently outperform classic and state-of-the-art static pruning methods.
Remarkably, DS-MBNet-M++ outperforms the \textit{sota} pruning methods EagleEye \cite{li2020eagleeye} and Meta-Pruning \cite{Liu2019MetaPruningML} by \textbf{2.3\%} and \textbf{2.6\%}. 

Third, DS-Net and DS-Net++ maintains superiority compared with powerful dynamic inference methods with varying depth, width or input resolution. For example, our DS-MBNet-M++ surpasses dynamic pruning method CG-Net \cite{hua2019channel} by \textbf{2.9\%}.

Fourth, DS-Net and DS-Net++ also outperforms its static counterparts. For instance, our DS-MBNet-S++ surpasses AutoSlim \cite{yu2019autoslim} and US-Net \cite{Yu2019UniversallySN} by \textbf{2.9\%} and \textbf{6.6\%}.

\begin{table}[t]
  \caption{Comparison of efficient inference methods with ResNet50 \cite{he2016deep} on ImageNet. \textcolor{DarkOrange3}{Orange}: NAS or pruning methods, \textcolor{DeepSkyBlue4}{Blue}: dynamic inference methods,  \textcolor{purple}{Red}: our method.
  }
  \label{tab:imagenet_resnet}
  \centering
\setlength{\tabcolsep}{9pt}
  \begin{tabular}{c|l|c|c}
    \toprule
    \multicolumn{2}{c|}{Method}      & MAdds & Top-1 \\
    \toprule
    \multirow{7}{*}{\makecell{3B\\MAdds}}
    & \textcolor{DarkOrange3}{SFP}~\pub{IJCAI18}~\cite{SoftFilterPruning}                                    & 2.9B & 75.1 \\
    & \textcolor{DarkOrange3}{ThiNet-70}~\pub{ICCV17}~\cite{luo2017thinet, liu2018RethinkingPruning}        & 2.9B & 75.8 \\
    & \textcolor{DarkOrange3}{MetaPruning 0.85}~\pub{ICCV19}~\cite{Liu2019MetaPruningML}                    & 3.0B & 76.2 \\
    & \textcolor{DarkOrange3}{AutoSlim}~\pub{arxiv19}~\cite{yu2019autoslim}                                & 3.0B & 76.0\\
    & \textcolor{DeepSkyBlue4}{ConvNet-AIG-50}~\pub{ECCV18}~\cite{veit2018AIG}                               & 3.1B & 76.2 \\
    & \textcolor{purple}{DS-ResNet-L (Ours)}                                              & 3.1B & 76.6\\
    & \textcolor{purple}{DS-ResNet-L++ (Ours)}                                              & 3.1B & \textbf{76.8}\\
    \midrule
    \multirow{9}{*}{\makecell{2B\\MAdds}}
    & ResNet-50 0.75$\times$ \cite{he2016deep}                                          & 2.3B & 74.9 \\
    & S-ResNet-50~\pub{ICLR19}~\cite{Yu2019SlimmableNN}                                              & 2.3B & 74.9 \\
    & \textcolor{DarkOrange3}{ThiNet-50}~\pub{ICCV17}~\cite{luo2017thinet, liu2018RethinkingPruning}        & 2.1B & 74.7 \\
    & \textcolor{DarkOrange3}{CP}~\pub{ICCV17}~\cite{He2017ChannelPF}                                       & 2.0B & 73.3 \\
    & \textcolor{DarkOrange3}{MetaPruning 0.75}~\pub{ICCV19}~\cite{Liu2019MetaPruningML}                    & 2.0B & 75.4 \\
    & \textcolor{DarkOrange3}{AutoSlim}~\pub{arxiv19}~\cite{yu2019autoslim}                                & 2.0B & 75.6\\
    & \textcolor{DeepSkyBlue4}{MSDNet}~\pub{ICLR18}~\cite{Huang2018MultiScaleDN}                             & 2.0B & 75.5 \\
    &  \textcolor{purple}{DS-ResNet-M (Ours)}                                             & 2.2B & 76.1\\
    &  \textcolor{purple}{DS-ResNet-M++ (Ours)}                                             & 2.3B & \textbf{76.4}\\
    \midrule
    \multirow{6}{*}{\makecell{1B\\MAdds}}
    & ResNet-50 0.5$\times$ \cite{he2016deep}                                           & 1.1B & 72.1 \\
    & \textcolor{DarkOrange3}{ThiNet-30}~\pub{ICCV17}~\cite{luo2017thinet, liu2018RethinkingPruning}       & 1.2B & 72.1 \\
    & \textcolor{DarkOrange3}{MetaPruning 0.5}~\pub{ICCV19}~\cite{Liu2019MetaPruningML}                     & 1.0B & 73.4 \\
    & \textcolor{DarkOrange3}{AutoSlim}~\pub{arxiv19}~\cite{yu2019autoslim}                                & 1.0B & 74.0\\
    & \textcolor{DeepSkyBlue4}{GFNet}~\pub{NeurIPS20}~\cite{wang2020glance}                                     & 1.2B & 73.8\\
    &  \textcolor{purple}{DS-ResNet-S (Ours)}                                             & 1.2B & 74.6\\
    &  \textcolor{purple}{DS-ResNet-S++ (Ours)}                                             & 1.2B & \textbf{75.0}\\
    \bottomrule
    \end{tabular}
\end{table}
\subsubsection{DS-ViT++}
We further evaluate our DS-Net++ with vision transformers on ImageNet. As illustrated in Fig. \ref{fig:imagenet}, DS-ViT++ outperforms other static models by a large gap in both accuracy and efficiency. Table \ref{tab:imagenet_t} provides a more detailed comparison. DS-ViT++ with different complexities consistently outperforms human designed ViT variants and other efficient inference method, i.e. NAS and dynamic inference methods. 

First, our DS-ViT++ models achieve \textbf{2.24}$\times$ and \textbf{3.62}$\times$ computation reduction over DeiT-S and ViT-S/16 with slight accuracy drop (-0.3\% and -0.6\%). Compared to statically scaled DeiT-Ti and DeiT-Ti+ model, DS-ViT-Ti++ and DS-ViT-S++ improves remarkably by \textbf{6.1\%} and \textbf{3.7\%}.

Second, the performance of our DS-ViT++ models surpasses static architecture optimization methods. For instance, DS-ViT-L++ outperforms two NAS models, AutoFormer and BossNAS-T, by \textbf{1.3\%} and \textbf{1.4\%} respectively.

Third, DS-ViT++ models also outperform other dynamic transformers. For example, DS-ViT-M++ improves over DynViT-DeiT and DVT-DeiT, the other two dynamic transformer based on ViT, by \textbf{2.3\%} and \textbf{0.6\%}, respectively.

\begin{figure*}
\centering
{
\centering
\hfill
     \begin{subfigure}[t]{0.3\textwidth}
         \centering
        \includegraphics[width=\linewidth]{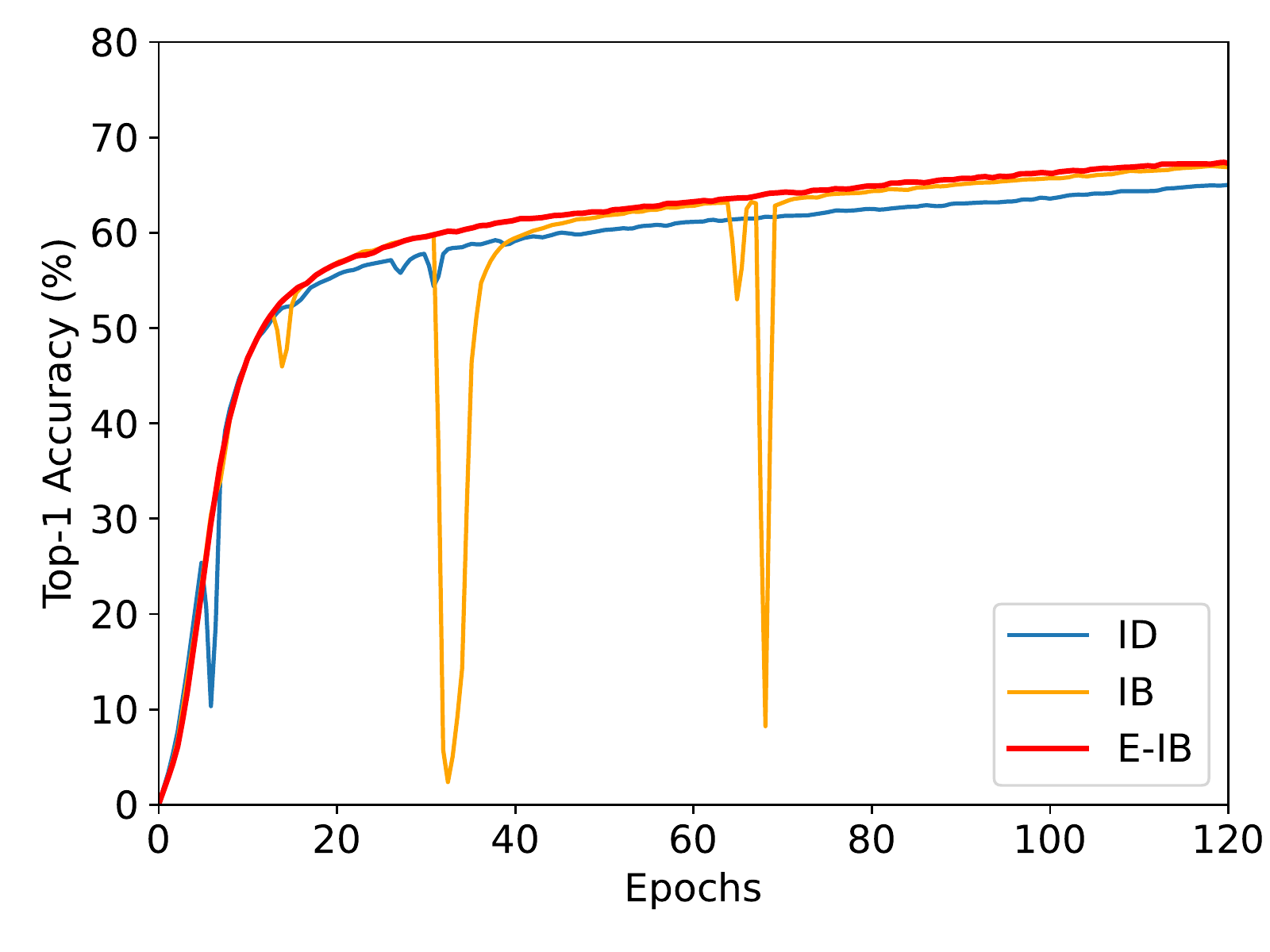}%
         \caption{Acc. of $\bm\psi^S$ in DS-MBNet}
     \end{subfigure}
     \begin{subfigure}[t]{0.3\textwidth}
         \centering
         \includegraphics[width=\linewidth]{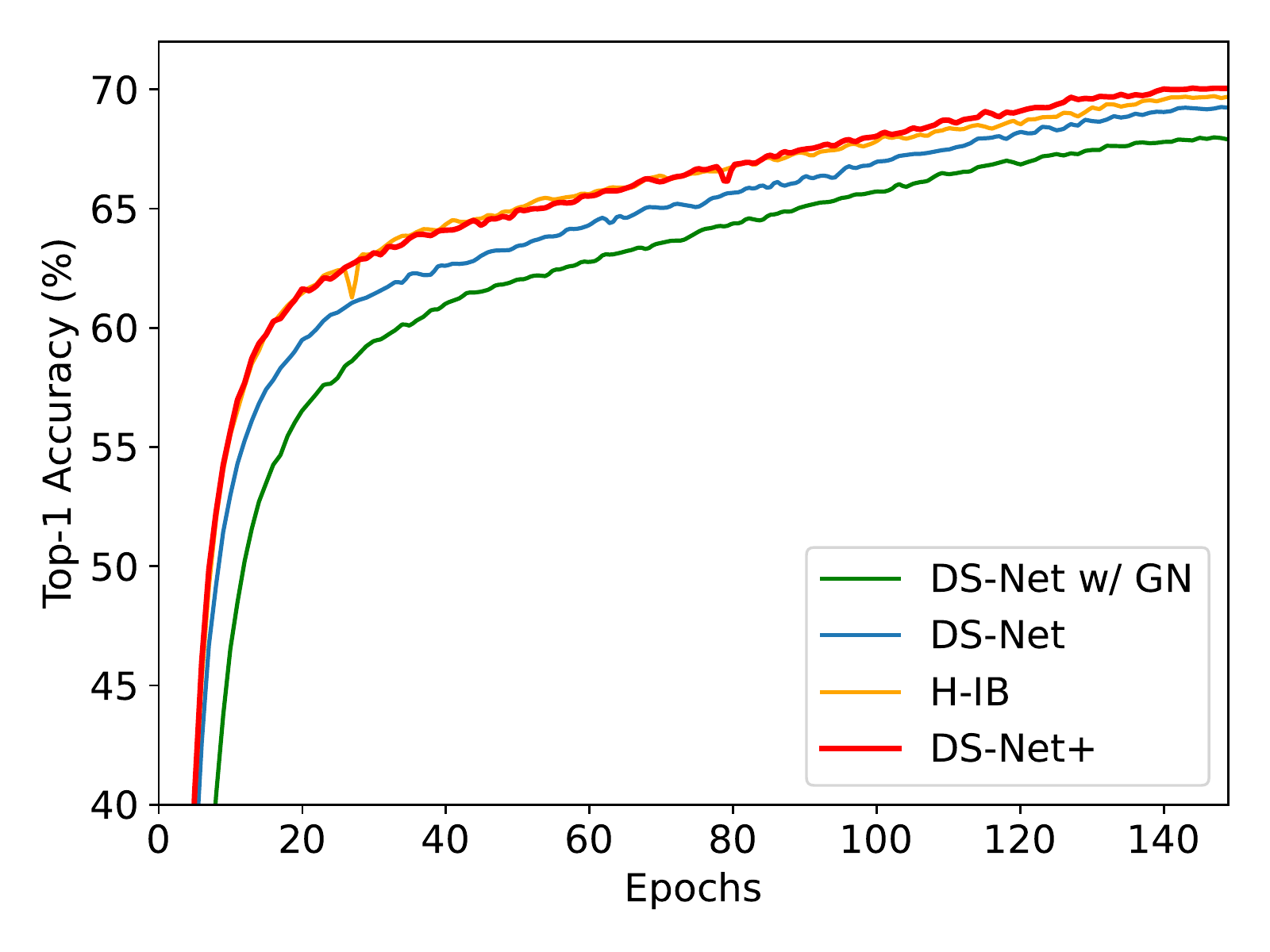}%
         \caption{Acc. of $\bm\psi^S$ in DS-MBNet+}
     \end{subfigure}
     \begin{subfigure}[t]{0.3\textwidth}
         \centering
         \includegraphics[width=\linewidth]{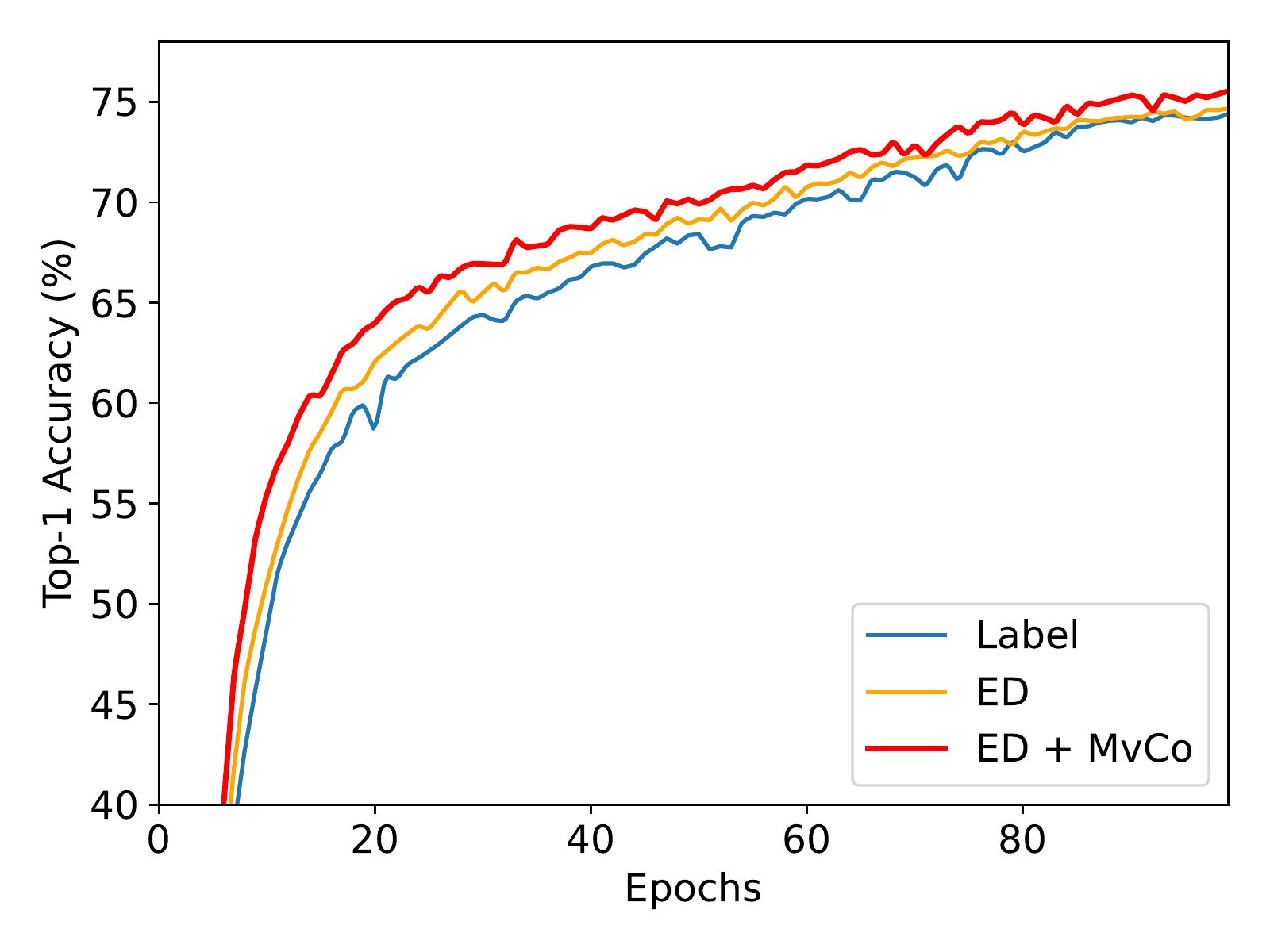}%
         \caption{Acc. of $\bm\psi^S$ in DS-ViT++}
     \end{subfigure}
\hfill
}
{\caption{Evaluation accuracy of the smallest sub-network $\bm\psi^S$ during supernet training with three different training schemes.}\label{fig:training_curve}}
\end{figure*}

\begin{table}[t]
  \caption{Comparison of state-of-the-art models based on ViT \cite{dosovitskiy2021vit} on ImageNet. \textcolor{DarkOrange3}{Orange}: NAS methods, \textcolor{DeepSkyBlue4}{Blue}: dynamic inference methods,  \textcolor{purple}{Red}: our method.
  }
  \label{tab:imagenet_t}
  \centering
  \begin{tabular}{c|l|c|c}
    \toprule
    \multicolumn{2}{c|}{Method}      & MAdds & Top-1 \\
    \toprule
    \multirow{6}{*}{\makecell{5B\\MAdds}}
    & ViT-S/16~\pub{ICLR21}~\cite{dosovitskiy2021vit}                                          & 4.7B & 78.8 \\
    & DeiT-S~\pub{ICML21}~\cite{touvron2020deit}                                           & 4.7B & 79.9 \\
    & T2T-ViT-14~\pub{ICCV21}~\cite{yuan2021tokens}    & 5.2B  & 81.5\\
    & \textcolor{DarkOrange3}{AutoFormer}~\pub{ICCV21}~\cite{chen2021autoformer} & 5.1B  & 81.7\\
    & \textcolor{DarkOrange3}{BossNAS-T $\bm\uparrow$}\pub{ICCV21}~\cite{li2021bossnas} & 5.7B & 81.6\\
    & \textcolor{purple}{DS-ViT-L++ (Ours)}                      &5.6B  & \textbf{83.0}\\
    \midrule
    \multirow{4}{*}{\makecell{4B\\MAdds}}
    & BoT50 + SE~\pub{CVPR21}~\cite{srinivas2021bottleneck} & 4.0B & 79.6 \\
    & Swin-T~\pub{ICCV21}~\cite{liu2021swin} & 4.5B & 81.3 \\
    & \textcolor{DeepSkyBlue4}{DynViT-DeiT 0.9$\times$}~\pub{arxiv21}~\cite{rao2021dynamicvit}   & 4.0B          & 79.8\\
    & \textcolor{DeepSkyBlue4}{DVT-DeiT}~\pub{arxiv21}~\cite{wang2021not}           &4.0B    &81.5\\
    & \textcolor{purple}{DS-ViT-M++ (Ours)}                      &4.1B  & \textbf{82.1}\\
    \midrule
    \multirow{5}{*}{\makecell{2B\\MAdds}}
    & DeiT-Ti+~\pub{ICML21}~\cite{touvron2020deit,d2021convit}  & 2.0B    & 75.9\\
    & ConViT~\pub{ICML21}~\cite{d2021convit} & 2.0B & 76.7 \\
    & T2T-ViT-12~\pub{ICCV21}~\cite{yuan2021tokens}    & 2.2B  & 76.5\\
    & PVT-tiny~\pub{ICCV21}~\cite{wang2021pyramid}   &1.9B   &75.1\\
    & \textcolor{DeepSkyBlue4}{DVT-DeiT}~\pub{arxiv21}~\cite{wang2021not}           &2.0B    &78.0\\
    &  \textcolor{purple}{DS-ViT-S++ (Ours)}                      &2.1B  & \textbf{79.6}\\
    \midrule
    \multirow{5}{*}{\makecell{1B\\MAdds}}
    & DeiT-Ti~\pub{ICML21}~\cite{touvron2020deit} & 1.1B & 72.1 \\
    & ConViT~\pub{ICML21}~\cite{d2021convit} & 1.0B & 73.1 \\
    & T2T-ViT-7~\pub{ICCV21}~\cite{yuan2021tokens} & 1.2B & 71.7 \\
    & \textcolor{DarkOrange3}{Autoformer}~\pub{ICCV21}~\cite{chen2021autoformer}      & 1.3B & 74.7\\
    &  \textcolor{purple}{DS-ViT-Ti++ (Ours)}                            &1.3B  & \textbf{78.2}\\
    \bottomrule
    \end{tabular}
\end{table}

\subsection{Transferability}
We first perform transfer learning in two settings with DS-ResNet on CIFAR-10: \textbf{a) DS-Net w/o gate transfer}: we transfer the supernet without routing gate to CIFAR-10 and retrain the dynamic gate. \textbf{b) DS-Net w/ gate transfer}: we first transfer the supernet then load the ImageNet trained gate and perform transfer leaning for the gate. We use gate transfer by default. The results along with the transfer learning results of the original ResNet and ViT models are shown in Table \ref{tab:cifar}. Gate transfer boosts the performance of DS-ResNet by 0.4\% on CIFAR-10, demonstrating the transferability of dynamic gate.

We further evaluate the transferability of DS-Net using DS-ResNet, DS-ResNet++ and DS-ViT++ on CIFAR-10 and CIFAR-100 datasets. Our transferred DS-ResNet++ models outperforms the original ResNet-50 by a large gap (1.1\% on CIFAR-10, 2.4\% on CIFAR-100) with about 2.5$\times$ computation reduction and even performs comparably to the larger ResNet-101 with 4.9$\times$ fewer computation complexity. Remarkably, DS-ViT++ outperforms the large ViT-B/16 model in both datasets with nearly 9.9$\times$ computation reduction. These strong results proves the superiority of DS-Net in transfer learning.

\subsection{Object Detection}
In this section, we evaluate and compare the performance of original MobileNet, DS-MBNet and DS-MBNet++ when used as the feature extractor in object detection with Feature Fusion Single Shot Multibox Detector(FSSD) \cite{li2017fssd}. We use the features from the 5-th, 11-th and 13-th depthwise convolution blocks (with the output stride of 8, 16, 32) of MobileNet for the detector. When using DS-MBNet and DS-MBNet++ as the backbone, all the features from dynamic source layers are projected to a fixed channel dimension by the feature transform module in FSSD \cite{li2017fssd}.

Results on VOC 2007 $\mathtt{test}$ set are given in Table \ref{tab:voc}. Compared to MobileNetV1, DS-MBNet-M and DS-MBNet-L with FSSD achieve 0.9 and 1.8 mAP improvement, while DS-MBNet-M++ and DS-MBNet-L++ further achieve 1.3 and 2.2 mAP improvements with 1.59$\times$ and 1.34$\times$ computation reduction respectively. These results demonstrates that our DS-Net and DS-Net++ remain their superiority after deployed as the backbone network in object detection task.
\begin{table}[t]
  \caption{Comparison of transfer learning performance on CIFAR-10 and CIFAR-100. MAdds are calculated on CIFAR-10 dataset. GT stands for gate transfer. 
  }
  \label{tab:cifar}
  \centering
  \footnotesize
  \begin{tabular}{l|c|c|c}
    \toprule
    Model                       & MAdds     & CIFAR-10 & CIFAR-100\\
    \midrule
    ResNet-50 \cite{he2016deep,Kornblith2018DoBI}& 4.1B      & 96.8  &84.6\\
    ResNet-101 \cite{he2016deep,Kornblith2018DoBI} & 7.8B         & 97.6&\textbf{87.0}\\
    \arrayrulecolor{lightgray}\hline\arrayrulecolor{black}
    DS-ResNet w/o GT               & 1.7B      & 97.4  &\na\\
    DS-ResNet                      & 1.6B      & 97.8  &86.8\\
    \arrayrulecolor{lightgray}\hline\arrayrulecolor{black}
    DS-ResNet++                   & 1.6B      & \textbf{97.9} & \textbf{87.0}\\
    \midrule
    ViT-B/16                    &  55.4B   & 98.1 & 87.1 \\
    DS-ViT++                   &  5.6B    & \textbf{98.7} & \textbf{90.1} \\
    \bottomrule
  \end{tabular}
\end{table}
\begin{table}[t]
  \caption{Performance comparision of DS-MBNet and MobileNet with FSSD on VOC object detection task.}
  \label{tab:voc}
  \centering
  \footnotesize
  \setlength{\tabcolsep}{14pt}
  \begin{tabular}{lcc}
    \toprule
    Model                   & MAdds   & mAP \\
    \midrule
    FSSD + MBNetV1 \cite{howard2017mobilenets,li2017fssd}         & 4.3B             & 71.9      \\
    \arrayrulecolor{lightgray}\hline\arrayrulecolor{black}
    FSSD + DS-MBNet-S      & 2.3B              & 70.7 \\
    FSSD + DS-MBNet-M      & 2.7B              & 72.8 \\
    FSSD + DS-MBNet-L      & 3.2B              & 73.7 \\
    \arrayrulecolor{lightgray}\hline\arrayrulecolor{black}
    FSSD + DS-MBNet-S++      &    2.3B              & 71.2\\
    FSSD + DS-MBNet-M++      &    2.7B              & 73.2\\
    FSSD + DS-MBNet-L++      &    3.2B              & \textbf{74.1}\\
    \bottomrule
  \end{tabular}
\end{table}
\begin{table}[t]
  
  \caption{Ablation analysis of proposed disentangled two-stage optimization scheme for dynamic networks.}
  \label{tab:ablation_twostage}
  \centering
  \begin{tabular}{l|ccc}
    \toprule
    Training scheme         & Model      & MAdds  & Top-1 Acc. \\
    \midrule
    \multirow{3}{*}{Single-Stage}
                            & smallest   & 133M  & 0.2 \\
                            & largest    & 565M  & 28.9 \\
                            & dynamic    & 255M  & 64.6\\
    \arrayrulecolor{lightgray}\hline\arrayrulecolor{black}
    \multirow{3}{*}{Two-Stage}               
                            & smallest   & 133M  & 64.3 \\
                            & largest    & 565M  & 74.1 \\
                            & dynamic    & 262M  & 69.5 \\
    \bottomrule
  \end{tabular}
\end{table}
\begin{table}[t]

  \caption{Ablation analysis of proposed training techniques for dynamic CNN supernets.}
  \label{tab:ablation_c}
  \centering
  \begin{tabular}{l|cc}
    \toprule
    Training scheme         & smallest  & largest \\
    \midrule
    In-place Distillation                                           & 66.5                  & 74.0\\
    + In-place Bootstrapping                                        & 68.1 (+1.6)           & 74.3 (+0.3)\\
    + Ensemble IB                                                   & 68.3 (+1.8)           & 74.6 (+0.6)\\
    + Re-SBN (\textcolor{purple}{\textit{DS-Net}})                   & 69.2 (+2.7)           & 75.9 (+1.9)\\
    \arrayrulecolor{lightgray}\hline\arrayrulecolor{black}
    + Hierarchical IB                                               & 69.8 (+3.3)           & 75.8 (+1.8)\\
    ~~~~[\textminus~Ensemble IB]                                    & 70.1 (+3.6)           & 76.3 (+2.3)\\
    + MvCo (\textcolor{purple}{\textit{DS-Net+}})       & 70.2 (+3.7)           & 76.1 (+2.1)\\
    + Compound Routing (\textcolor{purple}{\textit{DS-Net++}})   & \textbf{70.3 (+3.8)}  & \textbf{76.5 (+2.5)}\\
    \bottomrule
  \end{tabular}
\end{table}
\begin{table*}[t]

  \caption{Ablation analysis of proposed training techniques for dynamic transformer supernets. DS-ViT supernets are trained for 100 epochs, respectively. Results of DS-MBNet trained with some of the schemes (for 150 epochs) are also listed for comparison. (*: training can not converge.)}
  \label{tab:ablation_t}
  \centering
    \setlength{\tabcolsep}{9pt}
  \begin{tabular}{c|ccccc|cc|cc}
    \toprule
    & \multicolumn{5}{c|}{Training techniques} &\multicolumn{2}{c|}{DS-MBNet} &\multicolumn{2}{c}{DS-ViT} \\
    \midrule
    & Label & Sandwich & In-place Distill. & Distill. & MvCo & smallest & largest & smallest & largest \\
    \toprule
    \multirow{2}{*}{\makecell{Label\\Only}}
    &\cmark     & \xmark    & \xmark    & \xmark    & \xmark    & 61.5          & 71.9          & 70.2          & 77.6          \\
    &\cmark     & \cmark    & \xmark    & \xmark    & \xmark    & 64.3          & 74.1          & 74.4          & 81.4          \\
    \midrule
    \multirow{4}{*}{\makecell{In-place\\Distill}}
    &\xmark     & \cmark    & IB        & \xmarkg   & \xmarkg   & 69.2          & 75.9          & \pzo0.1*      & \pzo0.1*      \\ 
    &\xmark     & \cmark    & H-IB      & \xmarkg   & \cmark    & \textbf{70.2} & \textbf{76.1} & \pzo0.1*      & \pzo0.1*      \\ 
    &\cmarkg    & \cmark    & ID        & \xmarkg   & \xmarkg   & \na           & \na           & \pzo0.1*      & \pzo0.1*      \\ 
    &\cmarkg    & \cmark    & IB        & \xmarkg   & \xmarkg   & \na           & \na           & 68.1 (\textminus6.3)         & 74.3 (\textminus7.1)         \\  
    \midrule
    \multirow{2}{*}{\makecell{External\\Distill}}
    &\cmarkg    & \cmark    & \xmarkg   & \cmark    & \xmarkg   & \na           & \na           &  74.7 (+0.3)        & 81.2 (\textminus0.2)          \\ 
    &\cmarkg    & \cmark    & \xmarkg   & \cmark    & \cmark    & \na           & \na           & \textbf{75.5 (+1.1)} & \textbf{81.5 (+0.1)} \\
    \bottomrule
  \end{tabular}
\end{table*}
\begin{table}[t]
  \caption{Ablation analysis of routing gate.}
  
  \label{tab:slim_gate}
  \centering
  \footnotesize
  \begin{tabular}{lcccc}
    \toprule
    Model      & MBNet-S & MBNet-S++ & ResNet-S & ViT-S++ \\
    \midrule
    Supernet    & 69.3  & 70.3  & 73.4  & 79.3\\
    DS-Net      & \textbf{70.1}  & \textbf{70.8}  & \textbf{74.6}  & \textbf{79.6}\\
    \bottomrule
  \end{tabular}
\end{table}
\begin{table}[t]
  \caption{Ablation analysis of losses for gate training on ImageNet. Results in bold that use SGS loss achieve good performance-complexity trade-off. Fig. \ref{fig:ablation_gate} (a) gives a more intuitive illustration.}
  \label{tab:losses}
  \centering
  \footnotesize
  \begin{tabular}{c|c|c|c|c}
    \toprule
    Target      & Complexity        & SGS           & MAdds     & Top-1 Acc.\\
    \midrule
    \cmark  &                   &               & 3.6B      & 76.8  \\
                & \cmark        &               & 0.3B      & 66.2  \\
                &                   & \cmark~~ Give Up       & 1.5B              & 73.7 \\
                &                   & \cmark~~ Try Best   & \textbf{3.1B}      & \textbf{76.6}  \\
    \cmark  & \cmark        &               & 2.0B      & 75.0  \\
                & \cmark        & \cmark~~ Try Best   & \textbf{1.2B}      & \textbf{74.6}  \\
    \cmark  & \cmark        & \cmark~~ Try Best   & \textbf{2.2B}      & \textbf{76.1}  \\

    \bottomrule
  \end{tabular}
\end{table}

\subsection{Ablation study}
\subsubsection{Two-Stage Optimization versus Single-Stage Optimization.}
To analysis the effect of our proposed disentangled two-stage optimization scheme, we compare it with single-stage optimization schemes, where the gate and the supernet are jointly trained. In both of the training schemes, we do \textit{not} use extra training techniques (i.e. distillation, E-IB, etc.) to ensure an independent comparison.

As shown in Table \ref{tab:ablation_twostage}, the final dynamic network trained with two-stage optimization outperforms the one trained in single stage by \textbf{4.9\%}, demonstrating the effectiveness of two-stage optimization scheme. Moreover, the performance of the smallest and the largest sub-networks ($\bm\psi^S$ and $\bm\psi^L$) in dynamic supernet train with the two schemes are also shown in the table. $\bm\psi^S$ and $\bm\psi^L$ after single-stage training generalize poorly to the whole dataset, which supports our analysis in Section \ref{sec:twostage}.

\subsubsection{CNN Supernet Training}
\textit{In-place Bootstrapping \& Ensemble In-place Bootstrapping.}
We statistically analyse the effect of IB and E-IB technique using DS-MBNet with GroupNorm \cite{Wu2018GroupN}. We train a DS-MBNet supernet with three settings: original in-place distillation (ID), in-place bootstrapping (IB) and ensemble in-place bootstrapping (E-IB). As shown in Table \ref{tab:ablation_c}, $\bm\psi^S$ and $\bm\psi^L$ trained with IB surpassed the baseline by 1.6\% and 0.3\% respectively. With E-IB, the supernet improves by 1.8\% and 0.6\% on $\bm\psi^S$ and $\bm\psi^L$ comparing with ID. The evaluation accuracy progression curves of the smallest sub-network $\bm\psi^S$ during training with these three settings are illustrated in Fig.~\ref{fig:training_curve}~(a). The beginning stage of in-place distillation is unstable. Adopting EMA target improves the performance. However, there are a few sudden drops of accuracy in the middle of the training with EMA target. Though being able to recover in several epochs, the model may still be potentially harmed by those fluctuation. After adopting E-IB, the model converges to a higher final accuracy without any conspicuous fluctuations in the training process, demonstrating the effectiveness of our E-IB technique in stabilizing the training and boosting the overall performance of dynamic supernets. By equipping Split BatchNorm with Recalibration (Re-SBN), we have our strong basic DS-Net supernet achieving 69.2\% and 75.9\% accuracy with its smallest and largest sub-networks.

\textit{Hierarchical In-place Bootstrapping \& Multi-view Consistency.}
We further analyse our advanced training technique proposed for DS-Net++, \textit{i.e.} hierarchical in-place bootstrapping (H-IB) and multi-view consistency (MvCo), by training DS-MBNet supernet with Re-SBN using three different settings: \textbf{a)} using both E-IB and H-IB (+ H-IB), \textbf{b)} using H-IB only ([\textminus~E-IB]) and \textbf{c)} using H-IB + MvCo. When using H-IB, loss balancing factors in Eq. \eqref{eqn:h_ib_2} are set to $\lambda_\textit{IB} = \lambda_\textit{H-IB} = 0.5$. The H-IB loss is only calculated with the the last blocks' outputs following \cite{Romero2014FitNetsHF}. The results and training curve are shown in Table \ref{tab:ablation_c} and Fig.~\ref{fig:training_curve}~(b), respectively. Adding H-IB technique improves the performance of $\bm\psi^S$ in DS-MBNet supernet remarkably by 0.6\% with a minor drop (-0.1\%) on the largest sub-network $\bm\psi^L$. By disabling E-IB, the performance further surpasses the basic DS-Net supernet by 0.9\% and 0.4\% with $\bm\psi^S$ and $\bm\psi^L$, respectively. This significant gain over the DS-Net supernet confirms the effectiveness of the H-IB technique. We also apply multi-view consistency by default. However, ablation analysis shows that MvCo has no significant impact on CNN supernet. Nevertheless, we adopt MvCo to slightly improve the performance of $\bm\psi^S$, as it does not bring any extra training cost. Remarkably, by fully equipping these training techniques, the performance of $\bm\psi^S$ of DS-MBNet+ supernet improved by 1.0\% over DS-MBNet. By using compound routing space, accuracy of $\bm\psi^S$ and $\bm\psi^L$ is further improved by 0.1\% and 0.4\% in DS-MBNet++ supernet.

\subsubsection{Transformer Supernet Training}
We perform extensive experiments on training DS-ViT++ supernet with different training schemes, including in-place distillation and external distillation. When using label and distillation at the same time, the loss weight for each cross-entropy term is set to 0.5. All results of DS-ViT++ are abtained by training with pytorch automatic mixed precision for 100 epochs, and are shown in Table \ref{tab:ablation_t}. External distillation is performed with RegNetY-12GF \cite{radosavovic2020designing} as the teacher. Two label-only baselines with and without sandwich sampling are listed for reference. Results of DS-MBNet trained with some of the schemes (for 150 epochs) are also listed for comparison.
\label{sec:exp_vit_training}

\textit{Failure Cases of In-place Distillation.}
As shown in Table \ref{tab:ablation_t}, in-place distillation schemes fails to achieve reasonable results on DS-ViT++ supernet.
\textit{Firstly}, when trained with IB or H-IB + MvCo schemes designed for CNNs, \textit{i.e.} without label supervision for smaller sub-networks, the supernet can not converge.
\textit{Secondly}, when using ID + label, where smaller sub-networks are trained with hard label from the largest sub-network and the ground truth label, the training fails in the early stage, possibly because the training of smaller sub-networks easily collapse to uninformative solutions and stop the target network from learning useful information from the labels. 
\textit{Finally}, when disentangle the teacher's weight by with a EMA supernet, forming IB + label scheme, DS-ViT++ struggles to converge and eventually reaches a performance lower than label only baselines.

\textit{External Distillation \& Multi-view Consistency.}
As shown in Table \ref{tab:ablation_t} and Fig.~\ref{fig:training_curve}~(c), our proposed external distillation with multi-view consistency effectively improves the performance of DS-ViT++ supernet. By using hard distillation and label without MvCo, not ideally, DS-ViT improves by only 0.3\% in $\bm\psi^S$ and drops by 0.2\% in $\bm\psi^L$. By using MvCo, the performance of DS-ViT++ improves significantly by 0.8\% in $\bm\psi^S$ and 0.3\% in $\bm\psi^L$, achieving 1.1\% improvement over label only baseline in $\bm\psi^S$. These results demonstrate that MvCo scheme is very effective on vision transformer supernets.

\subsubsection{Gate Training}
\textit{Effect of Slimming Gate.} We analyse the improvement brought by routing gate by comparing the performance of DS-Net and its supernet. As shown in Table \ref{tab:slim_gate}, routing gate consistently boosts the performance of DS-Net and DS-Net++ models. Remarkably, routing gate improves the performance of DS-ResNet-S by 1.2\%, compared to sub-networks with similar sizes in its supernet.

\begin{figure}[t]
    \centering
    \begin{subfigure}[t]{0.45\linewidth}
    \includegraphics[width=\linewidth]{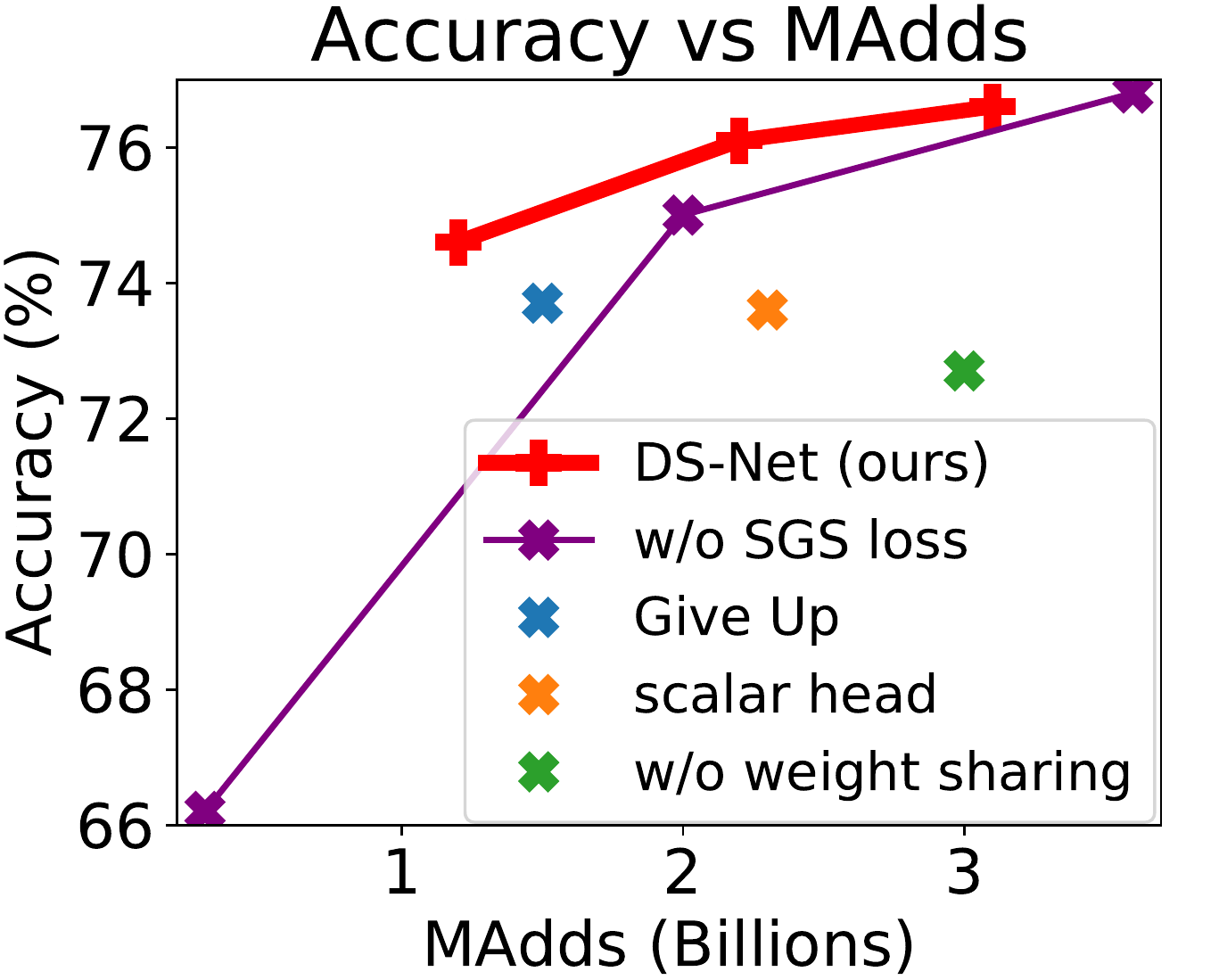}
    \caption{}
    \end{subfigure}~~%
    \begin{subfigure}[t]{0.55\linewidth}
    \includegraphics[width=\linewidth]{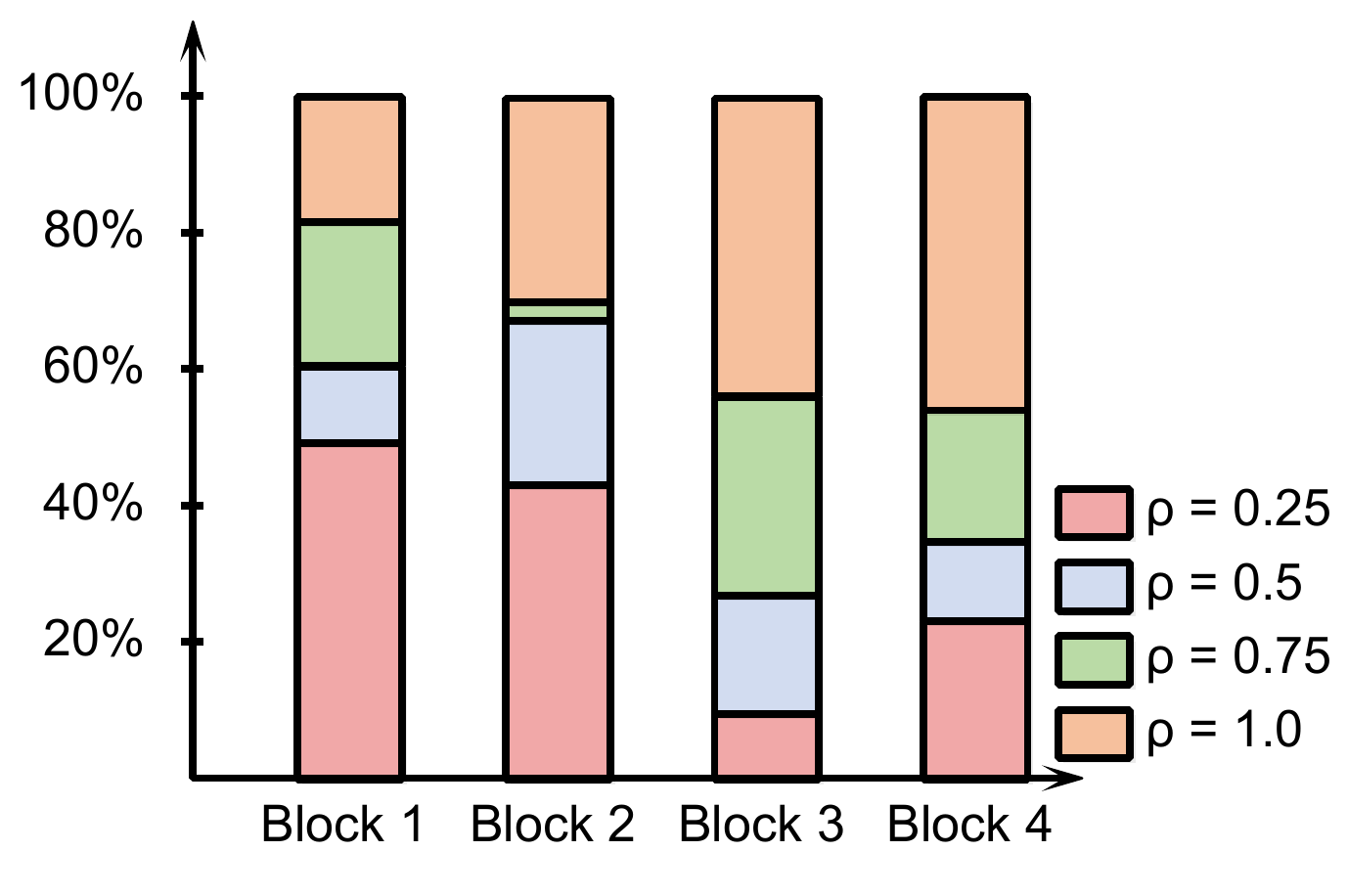}
    \caption{}
    \end{subfigure}
    \caption{\small\textbf{(a)} Illustration of accuracy \textit{vs.} complexity of models in Table \ref{tab:losses} and Table \ref{tab:gate_design}. \textbf{(b)} Gate distribution of DS-ResNet-M. The height of those colored blocks illustrate the partition of input samples that are routed to the sub-networks with respective routing signal $\bm\phi$. }
    \label{fig:ablation_gate}
\end{figure}

\textit{Effect of sandwich gate sparsification.}
To examine the impact of the three losses used in our gate training, \textit{i.e.} \emph{target loss} $\mathcal{L}_\textit{CLS}$, \emph{complexity penalty} $\mathcal{L}_\textit{CP}$ and \emph{SGS loss} $\mathcal{L}_\textit{SGS}$, we conduct extensive experiments with DS-ResNet on ImageNet, and summarize the results in Table \ref{tab:losses} and Fig. \ref{fig:ablation_gate}~(a). Firstly, as illustrated in Fig. \ref{fig:ablation_gate}~(a), models trained with SGS (\textcolor{Red3}{red} line) are more efficient than models trained without it (\textcolor{Purple4}{purple} line). Secondly, as shown in Table \ref{tab:losses}, with $\mathcal{L}_\textit{CLS}$, the model pursues better performance while ignoring computation cost; $\mathcal{L}_\textit{CP}$ pushes the model to be lightweight while ignoring the performance; $\mathcal{L}_\textit{SGS}$ itself can achieve a balanced complexity-accuracy trade-off by encouraging easy and hard samples to use small and large sub-networks, respectively.

\textit{SGS strategy.}\label{sec:SGS_strategy} Though we always want the easy samples to be routed to $\bm\psi^S$, there are two possible target definition for hard samples in SGS loss: \textbf{a)} \textit{Try Best}: Encourage the hard samples to pass through $\bm\psi^L$, even if they can not be correctly classified (\textit{i.e.} $\bm{y}_{\Phi}(\bm{x}_\textit{hard}) = \bm{e}^{|t|}$).
\textbf{b)} \textit{Give Up}: Push the hard samples to use $\bm\psi^S$ to save computation cost (\textit{i.e.} $\bm{y}_{\Phi}(\bm{x}_\textit{hard}) = \bm{e}^{1}$).
In both of the strategies, dependent samples are encouraged to use $\bm\psi^L$ (\textit{i.e.} $\bm{y}_{\Phi}(\bm{x}_\textit{dependent}) = \bm{e}^{|t|}$). The results for both of the strategies are shown in Table \ref{tab:losses} and Fig. \ref{fig:ablation_gate}~(a). As shown in the third and fourth lines in Table \ref{tab:losses}, \emph{Give Up} strategy lowers the computation complexity of the DS-ResNet but greatly harms the model performance. The models trained with \emph{Try Best} strategy (\textcolor{Red3}{red line} in Fig. \ref{fig:ablation_gate}~(a)) outperform the one trained with \emph{Give Up} strategy (\textcolor{DeepSkyBlue4}{blue} dot in Fig. \ref{fig:ablation_gate}~(a)) in terms of efficiency. This can be attribute to \emph{Give Up} strategy's optimization difficulty and the lack of samples that targeting on the largest path (dependent samples only 
account for about 10\% of the total training samples).
These results prove our \emph{Try Best} strategy is easier to optimize and can generalize better on validation set or new data.
\begin{table}[t]
  \caption{Ablation analysis of gate design on DS-ResNet.}
  \label{tab:gate_design}
  \centering
  \footnotesize
  \setlength{\tabcolsep}{8pt}
  \begin{tabular}{cccc}
    \toprule
    weight sharing   & routing head      & MAdds & Top-1 Acc. \\
    \midrule
    \cmark      & scalar        & 2.3B  & 73.6 \\
                    & one-hot       & 3.0B  & 72.7 \\
    \cmark      & one-hot       & 3.1B  & 76.6 \\
    \bottomrule
  \end{tabular}
\end{table}

\textit{Gate design.}\label{sec:gate_design}
\textbf{First}, to evaluate the effect of our weight-sharing double-headed gate design, we train a DS-ResNet without sharing the first fully-connected layer in $\Phi$ for comparison.
As shown in Table \ref{tab:gate_design} and Fig. \ref{fig:ablation_gate}~(a), the performance of DS-ResNet increase substantially (3.9\%) by applying the weight sharing design (\textcolor{SpringGreen4}{green} dot \textit{vs.} \textcolor{Red3}{red} line in Fig. \ref{fig:ablation_gate}~(a)). This might be attributed to overfitting of the routing head. As observed in our experiment, sharing the first fully-connected layer with attention head can greatly improve the generality.
\textbf{Second}, we also trained a DS-ResNet with \emph{scalar design}, where the routing head outputs a scalar ranges from 0 to 1, in contrast to our default \emph{one-hot design}.
As shown in Fig. \ref{fig:ablation_gate}~(a), the performance of \emph{scalar design} (\textcolor{DarkOrange3}{orange} dot) is much lower than the \emph{one-hot design} (\textcolor{Red3}{red} line), proving the superiority of \emph{one-hot design}.

\subsection{Gate Distribution Visualization}
To demonstrate the dynamic diversity of our DS-Net, we visualize the gate distribution of DS-ResNet over the validation set of ImageNet in Fig. \ref{fig:ablation_gate}~(b). In block 1 and 2, about half of the inputs are routed to $\bm\psi^S$ with 0.25 slimming ratio, while in higher level blocks, about half of the inputs are routed to $\bm\psi^L$. For all the gate, the slimming ratio choices are highly input-dependent, demonstrating the high dynamic diversity of our DS-Net.

\section{Conclusion} \label{sec:conclusion}
In this paper, we have proposed dynamic weight slicing scheme, achieving good hardware-efficiency by predictively slicing network parameters at test time with respect to different inputs. By using this scheme on CNNs filter numbers and more dimensions of CNNs and Transformers, we present DS-Net and DS-Net++. We propose a two-stage optimization scheme with IB, E-IB and SGS technique to optimize DS-Net and further propose H-IB and MvCo technique to optimize DS-Net++. We demonstrate that DS-Net and DS-Net++ can achieve 2-4$\times$ computation reduction and 1.62$\times$ real-world acceleration over CNNs and Transformers with minimal accuracy drops on ImageNet. Proved by experiments, DS-Net and DS-Net++ can surpass its static counterparts as well as state-of-the-art static and dynamic model compression methods on ImageNet by a large margin (up to 6.6\%). DS-Net and DS-Net++ generalize well on CIFAR-10, CIFAR-100 classification task and VOC object detection task.

\ifCLASSOPTIONcaptionsoff
  \newpage
\fi



\bibliographystyle{IEEEtran}
\bibliography{IEEEabrv,egbib}

\end{document}